%% file: main.tex
\documentclass{article}

\usepackage[preprint]{neurips_2026}

\usepackage{hyperref}
\usepackage[utf8]{inputenc} 
\usepackage[T1]{fontenc}    
\usepackage{url}            
\usepackage{booktabs}       
\usepackage{amsfonts}       
\usepackage{nicefrac}       
\usepackage{microtype}      
\usepackage{xcolor}         

\usepackage{wrapfig}
\usepackage{graphicx}
\usepackage{subcaption} 
\usepackage{enumitem} 

\usepackage{amsmath,amssymb,mathtools}
\usepackage{tikz}
\usetikzlibrary{
  arrows.meta,      
  calc,             
  patterns,         
  positioning,      
  fit,              
  decorations.pathreplacing, 
  matrix,           
}

\usepackage{bm}
\usepackage{multirow}
\usepackage{cleveref}
\usepackage{algorithm}
\usepackage[noend]{algpseudocode} 
\newcommand{\pluseq}{\mathrel{+}=}          

\title{Selective LoRA for Visual Tokens and Attention Heads}

\author{%
  Tiange Luo$^{1,2}$ \,
  Lajanugen Logeswaran$^{2}$ \,
  Jaekyeom Kim$^{2}$ \,
  Justin Johnson$^{1,\dagger}$ \,
  Honglak Lee$^{1,2,\dagger}$ \\
  University of Michigan$^{1}$ \quad
  LG AI Research$^{2}$ \quad
  $^{\dagger}$Equal advising
}

\begin{document}

\maketitle

\begin{abstract}
Low-rank adaptation (LoRA) is widely used for parameter-efficient fine-tuning, but its standard
all-token, all-head design ignores the heterogeneous structure of vision language model (VLM)
inputs. We introduce \emph{Image-LoRA}, a vision-oriented PEFT recipe that views LoRA as a
token-level residual update and applies this update only to visual tokens. Image-LoRA further
restricts adaptation to the value path of a compact subset of attention heads, selected using a
one-pass influence estimate from a rank-1 visual-token-only probe. This token-, head-, and value-selective design reduces trainable parameters and adapter-only training FLOPs while leaving the pure-text forward pass of the frozen backbone unchanged when no visual tokens are present. Across visual localization benchmarks with controlled text:image token ratios, Image-LoRA matches or closely approaches standard LoRA, while showing especially favorable trade-offs in image-token-heavy regimes. We further validate its generality on TextVQA and VideoQA, verify pure-text preservation on GSM8K, and show on ViLP that a stronger information bottleneck can yield gains over standard LoRA.
\end{abstract}

\section{Introduction}
\label{sec:introduction}

Low-rank adaptation (LoRA) \citep{hu2022lora} has become a standard recipe for parameter-efficient post-training because it keeps the base model frozen while learning a small trainable update that can approach full fine-tuning performance \citep{schulman2025lora}. Yet in vision--language models (VLMs), standard LoRA imposes only a weak structural prior. The same low-rank update is typically available to all token positions and attention heads, even though VLM inputs mix heterogeneous token spans---system text, user text, visual tokens, and output tokens---and prior work suggests that only some heads play a prominent role in visual routing, grounding, and pointing \citep{kang2025your,baek2025large}. This all-token, all-head treatment is potentially ineffective: different parts of the sequence need not be adapted in the same way, and not all heads need the same flexibility. It can also unnecessarily perturb the strong text behavior inherited from the frozen language backbone \citep{zhang2024wings,ratzlaff2025training,li2024multi,lu2025genieblue}, even when substantial language data are included during multimodal post-training \citep{bai2025qwen3,wang2025internvl3}. This raises a broader question: can we make adaptation more targeted---both in where it acts and in which heads it uses---to improve efficiency while preserving the text-only performance of the base model?
\begin{figure}[t]
    \centering
    \begin{subfigure}[t]{0.585\textwidth}
        \centering
        \includegraphics[width=\linewidth]{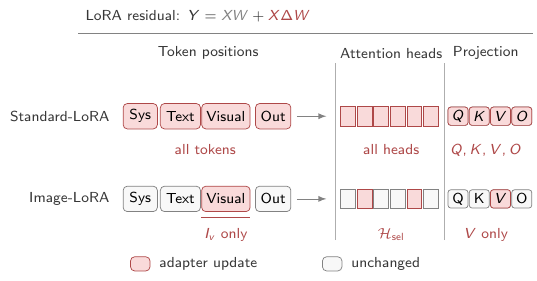}
        \label{fig:overview-left}
    \end{subfigure}\hfill
    \begin{subfigure}[t]{0.385\textwidth}
        \centering
        \includegraphics[width=\linewidth]{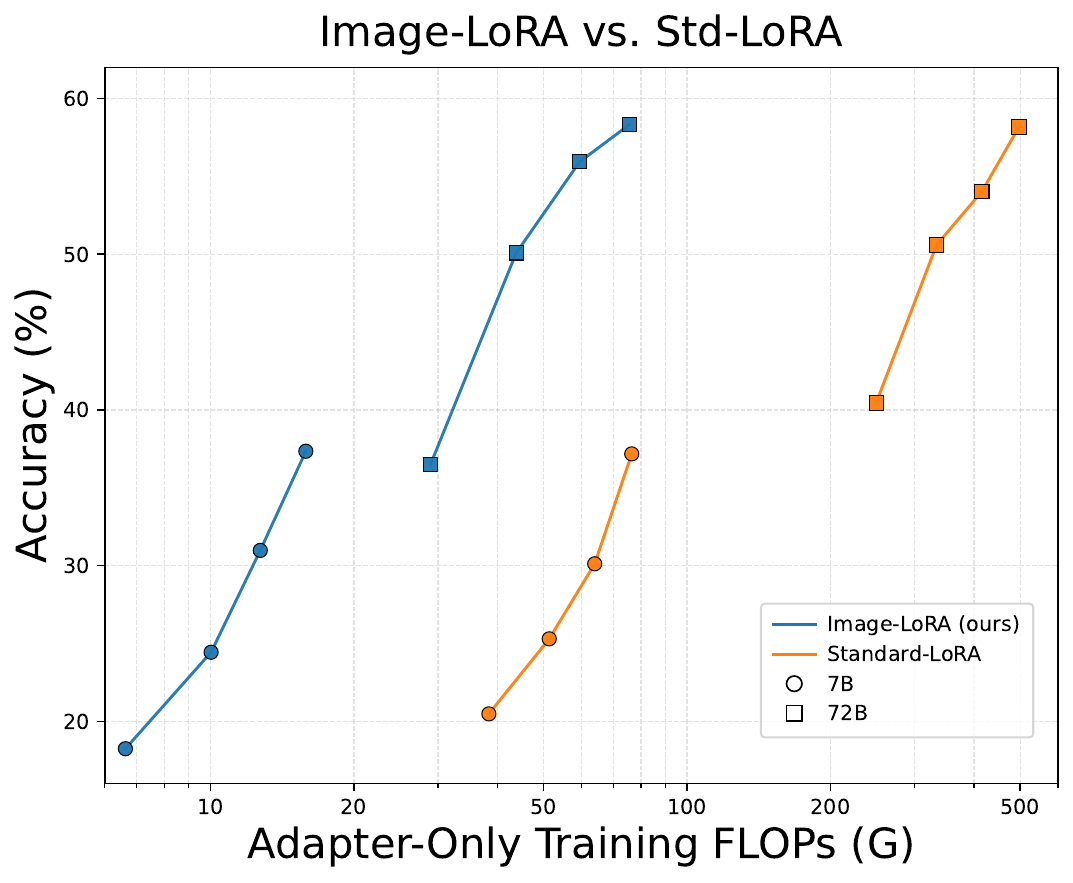}
        \label{fig:overview-right}
    \end{subfigure}
    \vspace{-0.15in}
    \caption{
\textbf{Left:} We interpret Standard-LoRA as applying residual updates to all tokens,
whereas Image-LoRA restricts these updates to the visual-token span, selected heads,
and value path. Standard-LoRA is commonly applied to the \(Q,V\) projections or to all attention projections \(Q,K,V,O\).
\textbf{Right:} Grounding-task accuracy versus adapter-only training FLOPs for
Qwen2.5-VL 7B and 72B. Points connect increasing input text:image token ratios
(\(1{:}2 \rightarrow 1{:}5\)), controlled by resizing images; FLOPs exclude the
frozen base model.
}
    \label{fig:overview}
    \vspace{-0.5em}
\end{figure}

Our starting point is a simple equivalent view of LoRA. For a linear projection with input sequence \(X\), frozen weight \(W\), and low-rank update \(\Delta W\), the adapted forward pass can be written as $Y = X(W+\Delta W) = XW + X\Delta W .$ 
Although LoRA is often described as merging \(\Delta W\) into the weight matrix, the term \(X\Delta W\) can also be viewed as a residual update applied to token representations. This view makes bottlenecking natural: in a VLM, the residual need not be applied to every token. We apply it only to the visual-token span \(\mathcal{I}_v\), so text-token positions receive no adapter residual at the injection site. For pure-text inputs, \(\mathcal{I}_v=\emptyset\), and the adapter is inactive; the model therefore exactly follows the frozen base network. This gives a direct preservation property while also tightening the adaptation bottleneck and reducing adapter-only training FLOPs by skipping prompt text and answer tokens. Importantly, visual-token-only adaptation can still affect downstream text generation, since text tokens can attend to adapted visual representations, and recent work suggests that information can be routed through visual-token channels in VLMs \citep{wei2025deepseek,cheng2025glyph}. \Cref{fig:overview} illustrates this token-selective contrast with standard LoRA.

Building on this token-level view, we propose \emph{Image-LoRA}, a vision-oriented PEFT recipe that is selective over tokens, heads, and projections. First, Image-LoRA adapts only visual tokens. Second, it adapts only a compact subset of attention heads. Rather than searching for such heads by enabling candidates one by one and recomputing validation losses, we attach a lightweight rank-1 visual-token-only probe to each head, run one forward--backward pass on the task loss, and rank heads by the sensitivity of the loss to their head-specific update. Since different layers may select different numbers of heads, we normalize the selected-head updates to stabilize layer-wise update magnitudes. Finally, within those selected heads, Image-LoRA adapts only the value path. This choice is well matched to our visual-token-only setting: from the perspective of a text token, visual keys help determine which image positions it attends to, while visual values help determine the content retrieved from those positions. Updating visual values therefore changes the visual evidence read by the language model, without directly modifying non-visual token representations at the adapter injection site. In our ablations, this value-only design outperforms alternative projection variants. Together, token, head, and value-path selectivity reduce trainable parameters and adapter-only training FLOPs.

We first evaluate Image-LoRA on visual localization benchmarks, including ScreenSpot-Pro \citep{li2025screenspot} and RefCOCO \citep{kazemzadeh-etal-2014-referitgame}, for Qwen2.5-VL-7B and Qwen2.5-VL-72B and controlled input text:image token ratios. These experiments cover regimes from text-heavy to image-token-heavy inputs, and show that Image-LoRA matches or closely approaches standard LoRA while using fewer trainable parameters and lower adapter-only training FLOPs, with especially favorable trade-offs when visual tokens dominate the context. \Cref{fig:overview} previews this accuracy--efficiency trend. To test whether our method generalizes beyond visual localization, we further evaluate it on the TextVQA~\citep{singh2019towards} and VideoQA~\citep{xiao2021next} datasets. We also extend our evaluation beyond Qwen2.5-VL by including LLaVA-NeXT-7B and InternVL3-8B.

Beyond visual-task accuracy, we also examine whether these stronger structural constraints can help adaptation. For pure-text inputs, \(\mathcal{I}_v=\emptyset\), so the adapter is inactive and the model exactly follows the frozen base network; we verify this behavior on GSM8K \citep{cobbe2021gsm8k}. More broadly, because updates are restricted to the visual tokens, selected attention heads, and value projections, Image-LoRA imposes a tighter information bottleneck than standard LoRA. Such a bottleneck may be beneficial in settings where broad adapters can exploit superficial textual cues instead of visual evidence. We probe this on ViLP \citep{luo2025probing}, a VQA benchmark designed to expose reliance on language priors, where Image-LoRA outperforms standard LoRA. Additional ablations validate the roles of visual-token-only adaptation, value-only projection selectivity, influence-based head selection, shared low-rank factors, and selection-size normalization. We discuss current limitations and open questions in Appendix~\ref{append:limitations}.

\section{Related Work}
\label{sec:related works}
\vspace{-0.1in}


\paragraph{PEFT for VLMs.}
A common approach to efficient fine-tuning of VLMs is to reduce the number of parameters to update, which is often called parameter-efficient fine-tuning (PEFT).
One important milestone in PEFT is reparameterizing model updates, notably represented by low-rank adaptation (LoRA) \citep{hu2022lora}, which has also been shown to be effective for VLMs \citep{sung2022vl,gao2023llama,zong2024mova, zanella2024low, dong2024internlm, wei2025moka}. 
Recently, Schulman and Thinking Machines Lab~\citeyearpar{schulman2025lora} examined the effects of applying LoRA either only to attention layers or to all layers, including MLP/MoE layers, a topic explored in prior works \citep{zaken2021bitfit,zhao2020masking,sung2021training,guo2020parameter,fu2023effectiveness, wang2025vision}. However, our work explores a more structured approach~\citep{hu2023vl, zhou2024empirical}—applying LoRA selectively to attention heads and to the visual token span only~\citep{wu2024reft}. Meanwhile, \citet{shuttleworth2024lora} reported findings that contradict \cite{schulman2025lora}, showing that LoRA can behave differently from full fine-tuning, with both studies focusing on pure-text benchmarks. In contrast, we focus on visual reasoning tasks, including grounding and VQA tasks.


\paragraph{Token selection and reduction for VLMs.}
Token selection and reduction have been applied to improve VLM inference efficiency.
Token pruning removes unimportant input tokens based on relevance criteria.
FastV~\citep{chen2024image} ranks visual tokens by attention scores, while SparseVLM~\citep{zhang2024sparsevlm} first filters relevant text tokens and then keeps the most attended visual tokens.
Token merging instead combines similar tokens to reduce sequence length.
PuMer~\citep{cao2023pumer} prunes low-saliency visual tokens and merges similar tokens within the same modality, and VisionZip~\citep{yang2025visionzip} merges non-dominant visual tokens based on attention scores.
On the other hand, we explore a different angle through token selection—investigating how constraining the set of input tokens can make the fine-tuning of VLMs (rather than inference) more efficient by applying Image-LoRA to the visual-token span only. This offers two benefits: (1) it saves FLOPs from both input and output text, and (2) such LoRA fine-tuning does not alter the pure-text reasoning behavior of VLMs.

\paragraph{Attention head selection for VLMs.}
Prior work shows that attention heads in LLMs and VLMs differ in their task importance.
\citet{michel2019sixteen} demonstrate redundancy among heads in language models and prune them by measuring loss sensitivity to head masking, while \cite{voita2019analyzing, zhang2023adalora} similarly assess head importance via backpropagated gradients and achieve comparable performance after pruning. Furthermore, \citet{khaki2025sparselora} accelerate LLM fine-tuning with an SVD-based sparsity estimator over the model's pre-trained weight matrix.
\citet{kang2025your, kang2025see} identify ``localization heads'' crucial for visual grounding and leverage their attention maps at inference, and \citet{zhang2025mllms} enhance perception by using the most attentive layer for test-time visual cropping.
This paper develops an efficient first-order approximation~\citep{molchanov2019importance} for head selection~\citep{zhang2025adaptive} as a simple complement to our proposed Image-LoRA, and shows that, when combined with the proposed layer-wise head-selection size normalization, our method performs well even with irregularly selected attention heads.

\input{sec/method}

\section{Experiments}
\label{sec:experiments}
We first compare \emph{Image-LoRA} and \emph{standard LoRA} on \textsc{Qwen2.5-VL-7B} and \textsc{Qwen2.5-VL-72B}, using ScreenSpot-Pro~\citep{li2025screenspot} and RefCOCO~\citep{kazemzadeh-etal-2014-referitgame} with controlled input text:image token ratios to study how the visual-token proportion affects performance. We then measure inference-time overhead, test whether pure-text reasoning is preserved on GSM8K~\citep{cobbe2021gsm8k} after visual-task fine-tuning, and broaden the evaluation to TextVQA and NExT-QA. Next, we examine the stronger information-bottleneck hypothesis on ViLP~\citep{luo2025probing} and verify transfer to \textsc{LLaVA-NeXT-7B} and \textsc{InternVL3-8B}. Finally, we present ablations on layer-wise normalization, projection choice, and other design variants.

\input{figs/main_table}

\subsection{Image‑LoRA vs. Standard LoRA}


\paragraph{Controlling the input text:image token ratio.}
This ratio matters for Image-LoRA because our method updates only visual tokens, so its cost and behavior should depend on how much of the input is visual. To study this cleanly, we keep the chat template fixed and change only the image resolution. In Qwen, larger images yield more visual tokens because the image is split into a regular patch grid, while the text portion stays nearly fixed at about \(850\) tokens, mostly from the system prompt. We preserve the original aspect ratio as much as possible, so this gives a clean way to vary the number of image tokens without changing the task itself; see Appendix~\ref{append:exp:main} for details. We use a grounding task with a one-line target, \texttt{point\_2d:[x, y]}, and we train using cross-entropy on that string. We report accuracy as the fraction of predicted \((x,y)\) points that fall inside the ground-truth box, along with trainable parameters and adapter-only training FLOPs excluding the frozen base model, on two visual localization benchmarks against standard LoRA. The setups are listed below.

\begin{itemize}[leftmargin=*]\itemsep3pt
  \item \textbf{ScreenSpot‑Pro~\citep{li2025screenspot} (UI grounding):} high‑resolution interface screenshots paired with language instructions and click/box annotations. Because screenshots are often $>$ \(3000{\times}2000\), we use \emph{image‑heavy} input text:image token ratios of \(1{:}2\), \(1{:}3\), \(1{:}4\), and \(1{:}5\). Average image token counts are in Table~\ref{tab:main}; example image resolutions per setting are in Table~\ref{tab:resolutions}. We use a fixed random split: \(1000\) examples for \emph{training} + \emph{head selection} and the remaining \(581\) for \emph{evaluation}. 
  \item \textbf{RefCOCO~\citep{kazemzadeh-etal-2014-referitgame} (referring expressions):} natural descriptions of target objects in COCO images with bounding boxes. As COCO images are relatively small (\(\approx 600{\times}640\)), we use ratios of \(1{:}1\) and \(1{:}1/2\) (input text:image). We sample \(2000\) \texttt{val} examples for \emph{training} + \emph{head selection} and evaluate on \(500\) examples from the \texttt{test} split with no overlap.
\end{itemize}

\begin{itemize}[leftmargin=*]\itemsep3pt
\item \textbf{Image‑LoRA} adapts only the \(V\) projection in attention layers and only on visual‑token spans, using rank \(r{=}8\) and scale \(\alpha{=}16\) unless otherwise noted. For each text:image ratio, we apply the proposed head selection strategy, choosing 28 out of 112 heads for \textsc{Qwen2.5-VL-7B} and 80 out of 640 for \textsc{Qwen2.5-VL-72B}.
Note that 28 and 80 correspond to the number of layers in these models, respectively.
\item The \textbf{Standard LoRA (Std-LoRA)} baseline adapts \(Q\) and \(V\) on all attention heads and all tokens with the same \(r\) and \(\alpha\). We choose \(Q\) and \(V\) because this is the most widely used configuration and is also the default setting in the official documentation.
We also compare with \(V\)-only Std-LoRA 
in terms of both FLOPs and performance. 
\end{itemize}



\paragraph{Token Ratio Experiment Results.}
\begin{wrapfigure}[16]{R}{0.30\linewidth}
  \centering
  \includegraphics[width=\linewidth]{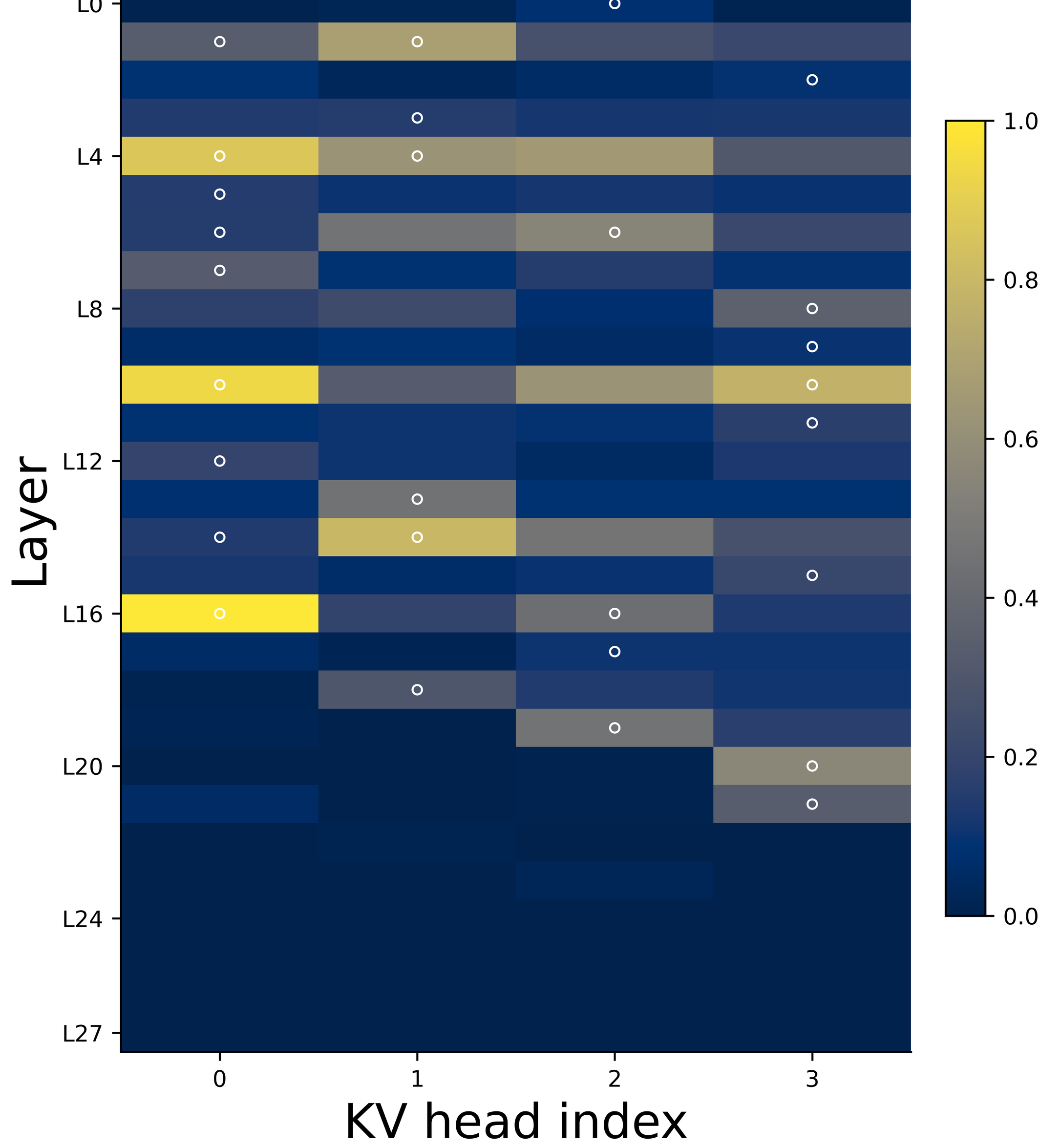}
  \vspace{-1.5em}
  \caption{Qwen2.5-VL-7B head selection at \(1{:}2\).}
  \label{fig:head_selection_example}
\end{wrapfigure}
Table~\ref{tab:main} groups settings by dataset and ratio, and compares \emph{Base (frozen)}, \emph{Std-LoRA (QV, all tokens, all 112 heads)}, and \emph{Image-LoRA (V, visual tokens only, 28 heads)} in accuracy, adapter FLOPs, and trainable parameters. Relative to \(QV\) Std-LoRA, Image-LoRA is substantially cheaper for two reasons: (1) it adapts fewer heads via our head-selection algorithm, which reduces trainable parameters, and (2) it computes updates only over the visual-token span, excluding prompt and answer tokens, which reduces compute. This yields the expected scaling: Image-LoRA FLOPs scale with the image-token count (so \(1{:}4 \approx 2\times 1{:}2\) and \(\approx 4\times 1{:}1\)), whereas Std-LoRA scales with the full sequence length (prompt + image + answer). On ScreenSpot-Pro, this translates into much lower adapter-only training FLOPs than \(QV\) Std-LoRA---for Qwen2.5-VL-7B, from 76.53/63.99/51.38/38.38G down to 15.85/12.72/10.03/6.63G across text:image ratios \(1{:}5\) to \(1{:}2\), and for Qwen2.5-VL-72B, from 496.9/415.5/333.7/249.2G down to 75.80/59.49/43.80/28.94G. In return, Image-LoRA matches or slightly exceeds \(QV\) Std-LoRA in the more image-heavy ScreenSpot-Pro settings (\(1{:}5\), \(1{:}4\)) and remains comparable on RefCOCO, but trails once the number of image tokens becomes smaller (\(1{:}3\), \(1{:}2\)). For reference, even compared with \(V\)-only Std-LoRA, Image-LoRA still saves approximately \(48\%\) adapter-only FLOPs on average for Qwen2.5-VL-7B on ScreenSpot-Pro; we defer the full 7B/72B comparison to Appendix~\ref{append:exp:ablation}, and Table~\ref{tab:abl-kv-path} shows that \(V\)-only Std-LoRA is consistently weaker than \(QV\) Std-LoRA.
Image-LoRA also uses far fewer parameters: roughly one-quarter of Std-LoRA for 7B models and one-fifth for 72B models. Figure~\ref{fig:head_selection_example} shows an example head selection for \textsc{Qwen2.5-VL-7B}, where selected heads span early through middle-late layers; we observe similar patterns for 72B model (more in Appendix~\ref{append:exp:main}).

\paragraph{Does Image-LoRA Add Inference-Time Cost?}
Because Image-LoRA applies the adapter only on the visual-token span \(\mathcal{I}_v\), it must dynamically identify those positions for each example and route the residual only there, making it a test-time adaptation rule than a static weight-merged update. In practice, the extra computation appears during prompt prefilling: for each layer \(\ell\), the adapter computes the shared projection for the \(T_v\) visual tokens and then the head-specific value updates for the \(N_{\ell}^{\text{chosen}}\) selected heads, giving a per-sample adapter forward cost of \(2T_v \sum_{\ell} (\dhid r + N_{\ell}^{\text{chosen}} r\dhead)\). During autoregressive decoding, newly generated tokens are text-only, so \(\mathcal{I}_v = \emptyset\) for those steps and no additional adapter FLOPs are incurred beyond prefilling. For Qwen2.5-VL-7B, this inference-time adapter cost is \(2.11/3.20/4.24/5.28\) GFLOPs per sample, and for Qwen2.5-VL-72B it is \(9.65/14.60/19.83/25.27\) GFLOPs, at text:image ratios \(1{:}2/1{:}3/1{:}4/1{:}5\), respectively.
\paragraph{Does Image-LoRA Affect Text-Only Reasoning?}
Because Image-LoRA adapts only the visual-token span, it does not change the model's pure-text computation path. To test whether text-only reasoning is preserved after visual-task adaptation, we fine-tune both Image-LoRA and Std-LoRA on ScreenSpot-Pro and then evaluate the resulting models on GSM8K~\citep{cobbe2021gsm8k}, which contains 1,319 math problems with no images. Qwen2.5-VL base achieves 25.55\%, and the ScreenSpot-Pro-adapted Image-LoRA model attains the same 25.55\%. Under the same ScreenSpot-Pro fine-tuning setup and text:image ratio, the corresponding Std-LoRA model drops to 24.79\%. This further validates that Image-LoRA preserves text-only reasoning under visual-task adaptation, whereas Std-LoRA, by adapting all tokens, can degrade it.

\paragraph{Does Image-LoRA Generalize Beyond Grounding?}
Beyond visual localization, we consider two complementary settings: TextVQA~\citep{singh2019towards}, which tests reasoning over text-rich images, and VideoQA~\citep{xiao2021next}, which tests temporally extended visual inputs with longer visual-token spans. For both settings, we adopt Qwen2.5-VL-7B. On TextVQA, with an average text:image token ratio of 1{:}1.15, Image-LoRA achieves 86.22, compared with 86.27 for Std-LoRA. For VideoQA, we represent 16 sampled frames as a single native video input and use an average of 2727.4 visual tokens per example, corresponding to an average text:image token ratio of 1{:}3.19. On the validation split, Std-LoRA reaches 78.0\%, and Image-LoRA improves slightly further to 78.4\%. Overall, this additional evaluation highlights that Image-LoRA is not specific to visual localization but extends naturally to image QA and temporally extended visual reasoning.

\paragraph{Does a Stronger Information Bottleneck Help?}
Image-LoRA imposes a tighter structural bottleneck than Std-LoRA: it restricts updates to the visual-token span, routes them through the value path, and applies them only to a selected subset of heads. This bottleneck may be useful in broader settings. Here, we instantiate this idea in a concrete case where the model must extract the answer from visual evidence rather than rely on superficial textual cues. We test this on ViLP~\citep{luo2025probing}, where each question is paired with three images: one ``follow'' image whose answer can be inferred from text alone and two ``against'' images whose correct answers contradict that textual prior and therefore require visual evidence. 
We fine-tune on ``follow'' and evaluate on the corresponding ``against'' images (the same questions, but with out-of-distribution answers). Image-LoRA outperforms Std-LoRA by 7.67\%, suggesting that the stronger bottleneck helps reduce shortcutting.

\paragraph{Does Image-LoRA Generalize to Other VLMs?}
We further evaluate Image-LoRA on LLaVA-NeXT-7B and InternVL3-8B. Because these models use discrete high-resolution tiling, the number of image tokens cannot be adjusted smoothly, so we keep each model's native image preprocessing and otherwise follow the Qwen setup. Since both base models achieve near-zero accuracy on ScreenSpot-Pro, we evaluate on RefCOCO instead. On LLaVA-NeXT-7B, Std-LoRA reaches \(91.2\%\) and Image-LoRA reaches \(92.0\%\), while reducing trainable parameters from \(3{,}407{,}936\) to \(786{,}454\). On InternVL3-8B, Std-LoRA reaches \(92.85\%\) and Image-LoRA reaches \(92.50\%\), while reducing trainable parameters from \(2{,}523{,}192\) to \(626{,}709\). Overall, these results suggest that Image-LoRA generalizes beyond Qwen while remaining substantially more parameter-efficient.

\subsection{Ablations on Image-LoRA}
\vspace{-0.1in}
\label{sec:ablations}

We ablate four Image-LoRA choices: selection-size normalization, KV-head budget, projection choice, and head-selection strategy. Unless noted otherwise, all runs use Qwen2.5-VL-7B on ScreenSpot-Pro with \(r{=}8\), \(\alpha{=}16\), and the same training setup as in the main experiments. Additional ablation studies appear in Appendix~\ref{append:exp:ablation}, including further analysis of our proposed head-selection procedure, different LoRA ranks, and how sharing $A$ affects the number of parameters.

\newsavebox{\ablheadselbox}
\newsavebox{\ablkvbudgetbox}
\newsavebox{\ablkvpathbox}
\newsavebox{\ablnormbox}

\begin{lrbox}{\ablnormbox}
\begin{minipage}[t]{0.485\textwidth}
  \centering
  \small
  \setlength{\tabcolsep}{4pt}
  \captionof{table}{\textbf{Selection-size normalization.} \(K_{\text{sel}}{=}28\), \(V\)-only, \(r{=}8\). Accuracy (\%).}
  \vspace{-0.25em}
  \label{tab:abl-norm}
  \begin{tabular}{l c c c c}
    \toprule
    Normalization & 1:5 & 1:4 & 1:3 & 1:2 \\
    \midrule
    None\ (\(\nu(S){=}1\))          & 34.60 & 29.43 & 23.41 & 17.73 \\
    Linear\ (\(\nu(S){=}1/S\))      & 35.97 & 27.71 & 24.44 & 17.21 \\
    Ours\ (\(\nu(S){=}1/\sqrt{S}\)) & \textbf{37.35} & \textbf{30.98} & \textbf{24.44} & \textbf{18.24} \\
    \bottomrule
  \end{tabular}
\end{minipage}
\end{lrbox}

\begin{lrbox}{\ablkvbudgetbox}
\begin{minipage}[t]{0.485\textwidth}
  \centering
  \small
  \setlength{\tabcolsep}{4pt}
  \captionof{table}{\textbf{KV-head budget.} \(K_{\text{sel}}\in\{1,7,28,112\}\). Accuracy (\%).}
  \vspace{-0.25em}
  \label{tab:abl-kv-budget}
  \begin{tabular}{l c c c c}
    \toprule
    Method & 1:5 & 1:4 & 1:3 & 1:2 \\
    \midrule
    Base (frozen) & 22.38 & 14.97 & 8.95 & 9.12 \\
    1 head        & 28.23 & 16.52 & 12.05 & 10.50 \\
    7 heads       & 30.81 & 24.27 & 18.42 & 15.15 \\
    28 heads (ours) & \textbf{37.35} & \textbf{30.98} & 24.44 & 18.24 \\
    112 heads     & 35.11 & 28.57 & \textbf{25.82} & \textbf{19.28} \\
    \bottomrule
  \end{tabular}
\end{minipage}
\end{lrbox}

\begin{lrbox}{\ablkvpathbox}
\begin{minipage}[t]{0.485\textwidth}
  \centering
  \small
  \setlength{\tabcolsep}{4pt}
  \captionof{table}{\textbf{Projection choices for Image-LoRA and Std-LoRA.} Accuracy (\%). }
  \vspace{-0.25em}
  \label{tab:abl-kv-path}
  \resizebox{\linewidth}{!}{%
  \begin{tabular}{l l c c c c}
    \toprule
    Method & Adapters & 1:5 & 1:4 & 1:3 & 1:2 \\
    \midrule
    Image-LoRA & $K$ only  & 29.60 & 24.10 & 19.10 & 13.77 \\
    Image-LoRA & \(V{+}K\) & 34.77 & 25.65 & 22.20 & 15.49 \\
    Image-LoRA & $V$ only  & \textbf{37.35} & \textbf{30.98} & 24.44 & 18.24 \\
    Image-LoRA & \(V{+}Q\) & 34.08 & 27.71 & 23.92 & 17.73 \\
    Image-LoRA & $QKVO$ & 34.42 & 29.78 & \textbf{25.82} & 20.48 \\
    \midrule
    Std-LoRA & $V$    & 35.28 & 29.60 & 24.78 & 19.45 \\
    Std-LoRA & \(V{+}Q\)   & 37.19 & 30.12 & 25.30 & 20.48 \\
    Std-LoRA & $QKVO$ & 34.77 & 28.74 & 25.30 & \textbf{21.51} \\
    \bottomrule
  \end{tabular}}
\end{minipage}
\end{lrbox}

\begin{lrbox}{\ablheadselbox}
\begin{minipage}[t]{0.485\textwidth}
  \centering
  \small
  \setlength{\tabcolsep}{4pt}
  \captionof{table}{\textbf{Head-selection methods.} All methods adapt \(K_{\text{sel}}{=}28\) heads. Accuracy (\%).}
  \vspace{-0.25em}
  \label{tab:abl-head-sel}
  \begin{tabular}{l c c c c}
    \toprule
    Method & 1:5 & 1:4 & 1:3 & 1:2 \\
    \midrule
    Global-Rand   & 34.42 & 28.40 & 21.86 & 16.87 \\
    PerLayer-Rand & 34.08 & 27.78 & 20.31 & 17.90 \\
    CorrMap       & 32.87 & 25.99 & 20.14 & 16.18 \\
    Ours          & \textbf{37.35} & \textbf{30.98} & \textbf{24.44} & \textbf{18.24} \\
    \bottomrule
  \end{tabular}
\end{minipage}
\end{lrbox}

\begin{table*}[!t]
  \centering
  \begin{minipage}[t]{0.485\textwidth}
    \usebox{\ablnormbox}\par\vspace{0.9em}
    \usebox{\ablkvpathbox}
  \end{minipage}\hfill
  \begin{minipage}[t]{0.485\textwidth}
    \usebox{\ablkvbudgetbox}\par\vspace{0.9em}
    \usebox{\ablheadselbox}
  \end{minipage}
\end{table*}


\paragraph{Selection-size Normalization.}
Because different layers select different numbers of heads, the raw increment $\sum_{h \in \mathcal{H}^\text{chosen}} \Delta V^{(h)}$ can scale with \(N^\text{chosen}\). We therefore use $s \;=\; \tfrac{\alpha}{r}\,\gamma\,\nu\!\big(N^\text{chosen}\big)$ and compare \textbf{None} (\(\nu(S){=}1\)) with \textbf{Linear} normalization (\(\nu(S){=}1/S\)). Table~\ref{tab:abl-norm} shows that both generally underperform our proposed \(\nu(S){=}1/\sqrt{S}\), with None tending to over-scale layers with many selected heads and Linear over-damping them.

\paragraph{Number of Chosen Heads.}
We vary the KV-head budget while fixing the one-shot influence selector. Qwen2.5-VL-7B has 28 layers and 112 KV heads in total, and we sweep $\{1, 7, 28, 112\}$ selected heads. Table~\ref{tab:abl-kv-budget} shows that accuracy generally improves with more heads, but not monotonically: at \(1{:}5\) and \(1{:}4\), selecting 28 heads outperforms using all 112.

\paragraph{Projection Choices.}
Table~\ref{tab:abl-kv-path} confirms \(V\)-only as the strongest overall Image-LoRA choice: it performs best in the more image-token-heavy settings (\(1{:}5\) and \(1{:}4\)), while \(K\)-only is weakest and both \(V{+}K\) and \(V{+}Q\) trail \(V\)-only. The denser \(QKVO\) variant beats \(V\)-only only at \(1{:}3\) and \(1{:}2\), but it requires over \(10.96\times\) more trainable parameters. For Std-LoRA, \(V{+}Q\) is strongest across most ratios, and \(QKVO\) helps only at \(1{:}2\). Overall, \(V\)-only offers the clearest efficiency--performance sweet spot when visual tokens dominate the input.

\paragraph{Head Selection Strategy.}
We compare our selector with three baselines: \emph{Global-Rand} (randomly sampling 28 heads across the model), \emph{PerLayer-Rand} (one head per layer, 28 total), and \emph{CorrMap} (a grounding-inspired heuristic based on image-token attention alignment with ground-truth regions, details in Appendix~\ref{append:exp:ablation}). Table~\ref{tab:abl-head-sel} shows that our method performs best across all ratios.

\section{Conclusion}
\label{sec:conclusion}
\vspace{-0.12in}
We introduced \emph{Image-LoRA}, a PEFT method for transformer-based VLMs that restricts adaptation to visual tokens and a subset of attention heads chosen by the proposed one-shot influence score, with layer normalization to stabilize updates. Across visual localization, TextVQA, and VideoQA benchmarks, Image-LoRA matches or closely approaches standard LoRA using fewer trainable parameters and lower adapter-only FLOPs, while preserving text-only reasoning.

\bibliography{main}
\bibliographystyle{icml2026}

\clearpage

\appendix

\input{sec/appendix}


\end{document}

%% file: sec/method.tex
\section{Method}
\label{sec:method-overview}
\vspace{-0.1in}

Image-LoRA follows directly from the residual view of LoRA introduced in \Cref{sec:introduction}. For an input sequence \(X\), standard LoRA adds the low-rank term \(X\Delta W\) on top of the frozen projection \(XW\). In a VLM, this residual need not be applied uniformly to every token position and every attention head. We therefore restrict adaptation to the visual-token span, a subset of heads, and the value path. This leaves the frozen model unchanged when no visual tokens are present while reducing trainable parameters and adapter-only training FLOPs.

Our method uses three simple forms of selectivity:
1. \textbf{Token selectivity (visual‑only):} We inject the adapter residual only on visual tokens, rather than on all tokens. This contrasts with standard LoRA-style PEFT, where the adapted projection is applied to every token position.
2. \textbf{Head selectivity (subset of attention heads):} We update only a compact subset of attention heads, selected using the one-pass influence-based procedure described in \Cref{sec:method-head-selection}.
3. \textbf{Projection selectivity (value‑only):} Rather than adapting multiple projections, we attach adapters only to the value ($V$) projection of the selected heads. This changes the visual content later consumed by attention without directly modifying non-visual token positions at the adapter injection site.

\subsection{Background}
\label{sec:method-background}

\newcommand{\dhid}{\ensuremath{d_{\mathrm{hidden}}}}
\newcommand{\dhead}{\ensuremath{d_{\mathrm{head}}}}

\paragraph{Transformer Multi-Head Attention.} For a sequence of length \(T\), let the input and output of a transformer layer be
\(X, Y \in \mathbb{R}^{T \times \dhid}\), where \(\dhid\) denotes the model width.
Multi-head attention can be written as
\[
Y = \mathrm{MultiHead}(X)
= \mathrm{Concat}(O^{(1)}, \dots, O^{(H)})\, W_O
\]
\[
\text{where } 
O^{(h)} = \mathrm{softmax}\!\left(
\frac{Q^{(h)} K^{(h)\top}}{\sqrt{d_k}} + Mask
\right) V^{(h)},
\]
\[
Q^{(h)} = X W_Q^{(h)},\ 
K^{(h)} = X W_K^{(h)},\
V^{(h)} = X W_V^{(h)}.
\]
Here \(H\) is the number of attention heads.
\(Mask \in \mathbb{R}^{T \times T}\) is the causal mask with \(Mask_{ij}=0\) if \(j \le i\) and \(Mask_{ij}=-\infty\) otherwise; the softmax is applied row-wise.
The query and key dimension and value dimension are denoted by \(d_k\) and \(d_v\), respectively (for simplicity, we assume $d_k = d_v = \dhid/H = 
\dhead$), and the learnable projection matrices are
$
W_{\{Q,K, V\}}^{(h)} \in \mathbb{R}^{\dhid \times \dhead}, 
W_O \in \mathbb{R}^{(H \times \dhead) \times \dhid}.
$

\paragraph{LoRA.}
LoRA is a parameter-efficient fine-tuning method that models updates to weight matrices as low-rank adaptations.
LoRA represents a linear map such as \(V=XW_V\) (superscript $(h)$ avoided for brevity) as a low-rank decomposition with $r \ll \min(\dhead, \dhid)$, 
\[
\Delta W_V = A B, \quad
A \in \mathbb{R}^{\dhid \times r},\ 
B \in \mathbb{R}^{r \times \dhead},\
\]
yielding the modified forward pass
\[
V = X (W_V + \Delta W_V)
   = X W_V + X A B \cdot s,
\qquad
s = \tfrac{\alpha}{r}\,\gamma,
\]
where \(\alpha\) is a scaling hyperparameter, \(\gamma\) is a learned scalar gate, and \(r\) is the adaptation rank.
The original weight \(W_V\) is frozen; only \((A,B,\gamma)\) are trained.
This makes the LoRA contribution \(X\Delta W_V \cdot s = XAB \cdot s\) explicit as an additive residual over token representations, which is the interpretation we use below.

\subsection{Image-LoRA}
\label{sec:method-ours}

\paragraph{Vision Tokens Only.}
The residual view above makes token selectivity immediate. Let the input prompt contain \(T\) tokens, of which \(T_v = |\mathcal{I}_v|\) correspond to vision tokens, where \(\mathcal{I}_v \subseteq \{1, \dots, T\}\) denotes the visual-token index set.
We instantiate the method on the value representations $V^{(h)}$. At non-visual positions the adapter contributes zero, and for pure-text inputs (\(\mathcal{I}_v = \emptyset\)) the layer exactly matches the frozen base model.
Later in the section we justify this choice and consider alternatives.
For a given attention head, we update the visual-token representations only:
\[
\tilde{v}^{(h)}_j = v^{(h)}_j + \Delta v^{(h)}_j, \qquad \text{where}
\Delta v^{(h)}_j := 0 \text{ for } j \notin \mathcal{I}_v
\]
and each $\Delta v^{(h)}_j$ is learned via LoRA as described next.

\paragraph{Selected Heads Only.}
Token selectivity alone still leaves every head adaptable. We therefore apply updates only to a subset of attention heads, chosen using the one-pass influence-based scoring procedure in \Cref{sec:method-head-selection}.
Let \(\mathcal{H}^{\text{chosen}} \subseteq \{1, \dots, H\}\) be the set of selected heads (see \Cref{sec:method-head-selection} for specific head selection criteria) and \(N^{\text{chosen}} = |\mathcal{H}^{\text{chosen}}|\).

Combining the two ideas above, we propose to learn parameters \(\Delta W_V^{(h)}\) such that
\[
\tilde{V}^{(h)} 
= X W_V^{(h)} 
+ \mathbb{I}[h \in \mathcal{H}^{\text{chosen}}] \, M\, X\, \Delta W_V^{(h)},
\]
where $M\in\{0,1\}^{T\times T}$ is a diagonal matrix with $M_{i,i} = \mathbb{I}[i \in \mathcal{I}_v]$.
This matrix formulation is provided for clarity; in practice, we parameterize \(\Delta W_V^{(h)}\) using a LoRA decomposition:
\[
\Delta W_V^{(h)} = A B^{(h)} s, 
\qquad
s = \tfrac{\alpha}{r}\,\gamma\,\nu(N^\text{chosen}),
\]
where \(A \in \mathbb{R}^{\dhid \times r}\) and \(B^{(h)} \in \mathbb{R}^{r \times \dhead}\) are learnable matrices, and the scaling factor \(s\) includes the normalization term \(\nu(N^\text{chosen})\), which we describe next.
In our implementation, \(A\) is shared across heads within a layer, while \(B^{(h)}\) is head-specific, motivated by two factors: (1) since \(\dhid \gg \dhead\)\footnote{For example, in Qwen2.5-VL-7B, \(\dhid=3584\), \(\dhead=128\).}, sharing \(A\) across a layer greatly reduces trainable parameters and adapter-only training FLOPs (detailed numbers for both 7B and 72B models in our ablation studies), and (2) head-specific parameters are required for the head-selection algorithm to compute per-head influence scores.

\paragraph{Selection-size Normalization.}
Standard LoRA activates \emph{all} heads per layer, so the per-layer update magnitude is tied to a fixed number of active heads.
With head selection, the number of chosen heads can vary across layers.
To prevent the summed increment from growing with the number of chosen heads, we additionally apply a \emph{selection-size normalization}:
\[
\nu(N^\text{chosen}) = \frac{1}{\sqrt{N^\text{chosen}}} .
\]
This design follows from a variance-preserving argument that ensures the variance of $\sum_{h \in \mathcal{H}^\text{chosen}} \Delta V^{(h)}$ remains invariant with respect to the number of chosen heads, assuming uncorrelated heads.
See \Cref{sec:layer_normalization} of the Appendix for more details.

\paragraph{Projection Selectivity.}
Finally, we restrict adaptation to the value path. Standard LoRA can be applied to any of the query $Q^{(h)}$, key $K^{(h)}$, value $V^{(h)}$, or output $O^{(h)}$ representations.
If we restrict the analysis to a single layer, updating the $Q^{(h)}$ or $O^{(h)}$ representations of vision tokens does not influence the output representations of text tokens. This is because
\[
o^{(h)}_i
= \mathrm{softmax}\!\left(
\frac{q^{(h)}_i K^{(h)\top}}{\sqrt{d_k}} + Mask_{i}
\right) V^{(h)},
\]
and thus $o^{(h)}_i$ depends only on the query for the token at $i$, together with the shared keys $K^{(h)}$ and values $V^{(h)}$. Consequently, modifying $q^{(h)}_j$ or $o^{(h)}_j$ for any vision token at $j \in \mathcal{I}_v$ has no effect on the output $o^{(h)}_i$ of any text token at $i \notin \mathcal{I}_v$.
By contrast, from the perspective of a text token, visual keys affect which image positions are attended to, while visual values determine the content retrieved from those positions. Updating visual values is therefore the most direct way to change the visual evidence consumed downstream without directly modifying non-visual token representations at the adapter injection site.
Based on our ablations, $V$-only emerges as the best overall choice for Image-LoRA: it is strongest in the more image-token-heavy regimes and consistently outperforms $K$-only, $V{+}K$, and $V{+}Q$, while the denser $QKVO$ variant helps only when image tokens are fewer and at substantially higher parameter cost.


\paragraph{Summary of Differences.}
Compared to the standard LoRA, Image-LoRA
(i) applies residual updates only to visual tokens, so pure-text inputs follow the frozen model exactly and training skips adapter computation over both prompt text and output text,
(ii) adapts only a selected subset of heads, with selection-size normalization to stabilize the magnitude of layer-wise updates, and
(iii) restricts adaptation to the value path while sharing \(A\) within a layer and keeping \(B^{(h)}\) head-specific, so the parameter count scales with \(|\mathcal{H}^\text{chosen}|\) instead of \(H\).
These design choices yield substantial parameter and adapter-only training FLOP reductions while still modulating the visual evidence available to later text tokens. Relative to the standard LoRA, our approach reduces the FLOPs by a factor of \(O(T_v/T \times |\mathcal{H}^{\text{chosen}}| / H)\).

\subsection{Head Selection}
\label{sec:method-head-selection}
\paragraph{First-order Head Importance Estimation.}
Our goal is to identify the attention heads $h$ in the model that would most effectively reduce the training loss when their value matrices are updated on the visual-token span.
Using a probe (validation) set $\mathcal{D}$ of size $S$, a na\"ive way to rank heads is to modify one head at a time, recompute the loss for all samples, and select the heads that yield the largest decrease.
If $N$ denotes the total number of candidate attention heads, this finite-difference procedure requires $O(NS)$ probe-loss evaluations and is prohibitively expensive.
Instead, we adopt a first-order Taylor approximation that scores \emph{all} heads from the same probe set using one backward pass per sample, reducing the probe cost to $O(S)$.
For parameters \(\theta\) and loss \(\ell\), the first-order Taylor expansion gives
\[
\Delta\ell = \ell(\theta+\delta\theta) - \ell(\theta)  \approx \langle \nabla_{\theta}\ell,\,\delta\theta\rangle,
\]
where higher-order terms \(O(\|\delta\theta\|^2)\) are neglected. Choosing the steepest direction of descent \(\delta\theta=-\nabla_{\theta}\ell\) predicts a loss change of
\[
\Delta\ell \approx -\,\|\nabla_{\theta}\ell\|_{F}^{2}
\]
Thus, $\|\nabla_{\theta}\ell\|_{F}^{2}$ can be used as a measure of the sensitivity of the loss to parameters $\theta$.

\paragraph{Head Importance.}
To measure the importance of each head, we adopt a rank-1 visual-token-only probe analogous to the Image-LoRA factorization described in \S\ref{sec:method-ours}.
Following standard practice, we initialize $A$ with Gaussian noise and set $B^{(h)}$ to zeros, with $\gamma=1$ and LoRA rank $r=1$ for efficiency.
The importance of head $h$ is then defined as
\[
I(h) = \|\nabla_{B^{(h)}} \ell\|_F^2
\]
which we refer to as the \emph{influence score}.
We accumulate this score over the probe set $\mathcal{D}$ to obtain one importance value per head.
Since $A$ is shared and frozen while $B^{(h)}$ is head-specific and trainable, the gradient magnitude on $B^{(h)}$ directly reflects how influential that head is.

\paragraph{Head Diversity.}
Besides importance, we consider a diversity factor, which we found beneficial in our experiments:
\[
F(h) = (\nabla_{B^{(h)}} \ell) \odot (\nabla_{B^{(h)}} \ell),
\]
where $\odot$ denotes the Hadamard (element-wise) product.
As with $I(h)$, we accumulate this feature over the same probe set and use it only to compare heads within a layer.

\paragraph{Head Selection Procedure.}
We next select a subset of heads that balances importance and diversity.  
To determine the number of heads to select from each layer, we first compute the aggregate importance of layer $L$ and convert it into a normalized weighting that guides head allocation:
\[
\Phi_L = \sum_{h \in L} I(h), \qquad
p_L = \frac{(\Phi_L)^{\tau}}{\sum_{L'} (\Phi_{L'})^{\tau}}
\]
where $\tau \ge 0$ is a temperature parameter controlling how strongly importance is concentrated across layers.
Given a total budget of $K_{\text{sel}}$ heads, we convert $p_L$ into an integer quota $k_L$ for each layer.
Within layer $L$, we first form a candidate pool consisting of the top $\lceil \rho k_L \rceil$ heads by importance (capped by the number of available heads in that layer), where $\rho \ge 1$ is a pool expansion factor, and then select $k_L$ heads using cosine farthest-first on row-normalized diversity features $F(h)$.
Appendix~\ref{app:head-selection-details} provides the exact quota-rounding, candidate-pool construction, and greedy selection details, while Algorithm~\ref{alg:ohi} summarizes the full probe-and-select pipeline.

%% file: figs/main_table.tex
\begin{table*}[t]
\centering
  \caption{\textbf{Main results across input text:image ratios.} FLOPs (G) denotes the total adapter-only training-time floating-point operations, measured in billions ($10^{9}$). The average fine-tuned parameter counts for Image-LoRA for 7B and 72B models are, respectively, 629,781 and 3,011,629. The parameter counts for Std-LoRA are 2,523,192 and 16,384,160, respectively.
  }
  \vspace{-0.1in}
  \label{tab:main}
\setlength{\tabcolsep}{5.2pt}
\resizebox{\textwidth}{!}{%
\begin{tabular}{l l c c c c c c c c c c c c}
\toprule
& & \multicolumn{6}{c}{\textsc{Qwen2.5-VL-7B}} &  \multicolumn{6}{c}{\textsc{Qwen2.5-VL-72B}} \\
    \cmidrule(lr){3-8}\cmidrule(lr){9-14}
& & \multicolumn{4}{c}{\textsc{ScreenSpot-Pro}} & \multicolumn{2}{c}{\textsc{RefCOCO}} & \multicolumn{4}{c}{\textsc{ScreenSpot-Pro}} & \multicolumn{2}{c}{\textsc{RefCOCO}}\\
    \cmidrule(lr){3-6}\cmidrule(lr){7-8}\cmidrule(lr){9-12}\cmidrule(lr){13-14}
\multicolumn{2}{c}{Text:Img Token Ratio} & 1:5 & 1:4 & 1:3 & 1:2 & 1:1 & 1:1/2 & 1:5 & 1:4 & 1:3 & 1:2 & 1:1 & 1:1/2 \\
\multicolumn{2}{c}{Avg \#Image Tokens} & 4,195 & 3,367 & 2,534 & 1,677 & 851 & 426 & 4,195 & 3,367 & 2,534 & 1,677 & 851 & 426 \\
\midrule
\multicolumn{1}{c}{\multirow{3}{*}{\shortstack{Acc. \\ (\%)}}}
& Base           & 22.38 & 14.97 & 8.95  & 9.12  & 90.80 & 83.00& 46.47 & 45.44 & 43.89 & 32.19 & 93.60 & 92.60 \\
& Std-LoRA       & 37.18 & 30.12 & \textbf{25.30} & \textbf{20.48} & 93.40 & \textbf{93.20}& 58.18 & 54.04 & \textbf{50.60} & \textbf{40.45} & 95.20 & \textbf{95.20} \\
& Image-LoRA     & \textbf{37.35} & \textbf{30.98} & 24.44 & 18.24 & \textbf{93.60} & 93.00& \textbf{58.35} & \textbf{55.94} & 50.09 & 36.49 & \textbf{95.80} & 94.80 \\
\midrule
\multirow{2}{*}{\shortstack{FLOPs \\ (G)}} 
& Std-LoRA       & 76.53 & 63.99 & 51.38 & 38.38 & 25.86 & 19.42& 496.9 & 415.5 & 333.7 & 249.2 & 167.9 & 12.61 \\
& Image-LoRA     & \textbf{15.85} & \textbf{12.72} & \textbf{10.03} & \textbf{6.63} & \textbf{3.36} & \textbf{1.61} & \textbf{75.80} & \textbf{59.49} & \textbf{43.80} & \textbf{28.94} & \textbf{15.05} & \textbf{7.54} \\
\bottomrule
\end{tabular}
}
  \vspace{-2pt}
\end{table*}

%% file: sec/appendix.tex
\input{sec/method_appendix}


\section{Experimental Details}
\label{append:exp}
\subsection{Main Experiment}
\label{append:exp:main}
\paragraph{Training details.}
Image-LoRA adapters are injected \emph{pre‑RoPE} and \emph{before cache updates}. 
Unless noted, rank \(r{=}8\), \(\alpha{=}16\), no dropout, and LoRA parameters are in \(\mathrm{fp32}\). \emph{Standard LoRA} follows the \emph{QV} practice recommended in Qwen’s training docs\footnote{\url{https://qwen.readthedocs.io/en/latest/training/llama_factory.html}}, i.e., it adapts the \(q\)- and \(v\)-projections on \emph{all} tokens and \emph{all} heads (no head selection, no visual‑token only restriction), using the same rank \(r\) and scale \(\alpha\). We train for 5 epochs with AdamW, learning rate \(5\times 10^{-4}\), batch size 8, and cosine lr scheduler. 
For Image‑LoRA we share one \(A^{(L)}\in\mathbb{R}^{H\times r}\) per layer and fit head‑specific \(B^{(L)}_{v,j}\in\mathbb{R}^{r\times d_{head}}\) only for selected KV heads; at inference we freeze \(\{A,B,\gamma\}\) and apply the same visual‑slice update so non‑visual tokens incur no additional compute. 
Head selection uses a fixed global budget of $K_{sel}=28$ KV heads across the 7B model. Since \textsc{Qwen2.5‑VL‑7B} has \(28\) layers and grouped‑query attention with \(g{=}7\) (\(H_q{=}28\Rightarrow H_{kv}{=}4\) per layer), selecting 28 KV heads amounts to a \(\tfrac{28}{28\times 4}=1/4\) fraction of all KV heads. For 72B model, we have $K_{sel}=80$, where \textsc{Qwen2.5‑VL‑72B} has \(80\) layers and grouped‑query attention with \(g{=}8\) (\(H_q{=}64\Rightarrow H_{kv}{=}8\) per layer), selecting 80 KV heads amounts to a \(\tfrac{80}{80\times 8}=1/8\) fraction of all KV heads. The data used for head selection are always the same as the training set in each dataset (no risk of evaluation set leakage).

\paragraph{Controlling the input-text:image token ratio details.}
We control the ratio by sweeping the processed image resolution while keeping the chat template fixed. The visual span \(\mathcal{I}_v\) is delimited by \texttt{<|vision\_start|>}\ldots\texttt{<|vision\_end|>}. 
As Qwen models tokenize images on a regular patch grid, the number of visual tokens, \(T_v\), grows monotonically with the processed area. For \textsc{Qwen2.5-VL} with \texttt{patch\_size}$=14$ and \texttt{merge\_size}$=2$, the effective stride is \(s=28\) px and $T_v \approx \Big\lfloor \tfrac{H}{s} \Big\rfloor \Big\lfloor \tfrac{W}{s} \Big\rfloor.$ The text side effectively contains about (\(T_{\text{text}}\!\approx\!850\) tokens, mostly from system prompts, while each per-sample query remains short. 
Given a target ratio \(\rho := T_{\text{text}}/T_v\), we set \(T_v^\star \approx T_{\text{text}}/\rho\) and binary-search the processed longest side—bounded by \texttt{min\_pixels}/\texttt{max\_pixels} and quantized to multiples of \(s\)—until \(|T_v - T_v^\star|\le \varepsilon\). The whole process only resamples the image while preserving its original aspect ratio as much as possible. It cannot achieve a perfectly fixed text:image token ratio for every sample since the resolution must be a multiple of 28, but it yields ratios that are very close across samples, enabling consistent sweeps from text-heavy to image-heavy supervision.

\begin{table}[!t]
  \centering
  \caption{\textbf{Example image resolutions for each input text:image token ratio.} The average image token counts (\textbf{\#Image Tokens}) correspond to Table~\ref{tab:main}. We provide two example image resolutions for reference.}
  \label{tab:resolutions}
  \setlength{\tabcolsep}{4pt}
  \begin{tabular}{l c c c}
    \toprule
    Ratio & \#Image & Example Res. \#1 & Example Res. \#2 \\
    & Tokens & (W$\times$H) & (W$\times$H) \\
    \midrule
    \multicolumn{4}{l}{ScreenSpot-Pro} \\
    \addlinespace[1pt]
    1:5 & 4{,}195 & 3472$\times$952 & 2240$\times$1456 \\
    1:4 & 3{,}367 & 3108$\times$840 & 2072$\times$1260 \\
    1:3 & 2{,}534 & 2660$\times$756 & 1764$\times$1092 \\
    1:2 & 1{,}677 & 2184$\times$616 & 1540$\times$840 \\
    \midrule
    \multicolumn{4}{l}{RefCOCO} \\
    \addlinespace[1pt]
    1:1 & 851 & 1036$\times$644 & 952$\times$700 \\
    2:1 & 426 & 700$\times$476 & 588$\times$560 \\
    \bottomrule
  \end{tabular}%
  \vspace{-2pt}
\end{table}


\subsection{Ablation Studies}
\label{append:exp:ablation}

\paragraph{Head Selection Procedure More Analysis}


We study two orthogonal knobs in our head--selection method:
(i) the \emph{layerwise budget temperature}~$\tau$, which distributes
the total head budget across layers (see Section~\ref{sec:method-head-selection}
and Appendix~\ref{app:head-selection-details}), and
(ii) the \emph{within-layer} selection rule (diversity-aware vs.\ importance-only),
controlled by the pool expansion factor~$\rho$ (Appendix~\ref{app:head-selection-details}).
Unless stated otherwise, we select $K_{\text{sel}} = 28$ KV heads out of
$H_{\text{tot}} = 112$ in Qwen2.5-VL-7B (28 layers~$\times$~4 KV heads per layer).

\emph{Layerwise budgeting via $\tau$.}
Recall that we aggregate per-layer importance
$\Phi_L = \sum_{h \in L} I(h)$ and form allocation weights
$p_L \propto \Phi_L^{\tau}$, which are then converted into integer quotas
$k_L$ under a global budget $K_{\text{sel}}$ (see Appendix~\ref{app:head-selection-details}
for the exact rounding and correction scheme).
Intuitively,
$\tau = 0$ yields an \emph{approximately uniform} allocation across layers,
$\tau = 1$ makes the budget \emph{proportional to mass}~$\Phi_L$,
and $0 < \tau < 1$ \emph{compresses} differences in $\Phi_L$ while still favoring
high-mass layers.
We use $\tau = 0.5$ as our default and additionally compare $\tau \in \{0,1\}$. We list some head selection examples in Figure~\ref{fig:headselection_different_tau}.
For Qwen2.5-VL-7B with $L = 28$ layers and $K_{\text{sel}} = 28$,
the $\tau = 0$ setting degenerates to $k_L = 1$ for every layer (up to rounding),
i.e., selecting exactly one KV head per layer—the one with the highest importance score.

\input{figs/tab_appen_headselection}

\input{figs/headselection_different_tau}
\input{figs/headselection_different_rho}

\emph{Diversity vs.\ importance-only.}
Given a layerwise budget $\{k_L\}$, we ablate the within-layer selection rule.
The \emph{importance-only} baseline simply picks, in each layer~$L$, the
top-$k_L$ heads by importance score $I(h)$.
The \emph{diversity-aware} variant first forms a pool
of size $k'_L = \min\!\big(H^{(L)}_{\text{kv}},\,\lceil \rho\,k_L\rceil\big)$
by taking the top-$k'_L$ heads in $L$ by $I(h)$, and then applies
cosine farthest-first selection on the diversity features $F(h)$
(Section~\ref{sec:method-head-selection} and
Appendix~\ref{app:head-selection-details}) to obtain a diverse subset
of size $k_L$.
The expansion factor $\rho \ge 1$ controls the trade-off:
$\rho = 1$ reduces exactly to pure top-$k_L$ by importance,
while larger $\rho$ allows more diversity at a small cost in $I(h)$.
In our main experiments we fix $\tau = 0.5$ and compare
(i) the importance-only baseline ($\rho = 1$) and
(ii) diversity-aware selection with $\rho \in \{2.0, 3.0\}$, where $\rho = 2.0$ is our default setting. We list some head selection examples in Figure~\ref{fig:headselection_different_rho}.

The results in Table~\ref{tab:abl-tau-diversity} validate our choices for the main experiments, namely setting $\tau = 0.5$ and $\rho = 2.0$.


\paragraph{More LoRA Ranks}
We report the results for different LoRA ranks $r=\{2,8,32\}$ in Table~\ref{tab:abl-lora-rank}.

\begin{table}[t]
  \centering
  \caption{\textbf{LoRA rank ablation on Qwen2.5-VL-7B.} 
  All methods adapt \(K_{\text{sel}}{=}28\) heads. Entries are accuracy (\%).}
  \vspace{-0.1in}
  \label{tab:abl-lora-rank}
  \begin{tabular}{l c c c c}
    \toprule
    LoRA rank & 1:5 & 1:4 & 1:3 & 1:2 \\
    \midrule
    $r = 2$   & 33.91 & 26.16 & 23.58 & 16.52 \\
    $r = 8$   & 37.35 & 30.98 & 24.44 & 18.24 \\
    $r = 32$  & 38.55 & 30.29 & 25.65 & 18.93 \\
    \bottomrule
  \end{tabular}
\end{table}

\paragraph{Effect of sharing \(A\).}
Sharing \(A\) across heads within each layer reduces adapter-only training FLOPs and parameters on ScreenSpot-Pro across a range of text:image token ratios. With Qwen2.5-VL-7B, the average adapter-only FLOPs drop by \(\approx 21.1\%\) and the parameter count by a similar fraction (\(\approx 172{,}032\) fewer). With Qwen2.5-VL-72B, the average FLOPs reduction is \(\approx 26.7\%\) with a comparable parameter decrease of \(\approx 26.6\%\) (\(\approx 1{,}064{,}960\) fewer).

\paragraph{More Parameters \& FLOPs information}
In the results analysis of Table~\ref{tab:main}, we report the FLOPs savings for the Qwen-2.5-VL-7B model when comparing V-only Std-LoRA with our Image-LoRA. Here, we provide detailed information on the different Std-LoRA key configurations for both the 7B and 72B models, including trainable parameter counts and adapter-only training FLOPs. All methods adopt a LoRA rank of 8. The results are shown in Tables~\ref{tab:sreenspot7b_lora_param_flops} and \ref{tab:sreenspot72b_lora_param_flops}. Our Image-LoRA is applied solely to the value path $V$ and to visual tokens, resulting in significantly lower adapter-only training FLOPs.

\paragraph{Correlation-based Head Selection}
For each image–instruction pair, we isolate the vision‑token span and the set of query tokens, extract the attention tensor for every layer $l$ and head $h$, average it over query positions to obtain a per‑head image‑attention map $a^{(i)}_{l,h}\in\mathbb{R}^{V}$ (with $V=g_h g_w$ vision tokens, each corresponding to a spatial cell in the vision tokenizer/encoder’s grid so that $a^{(i)}_{l,h}$ can be interpreted as a coarse attention pattern over the image), and compute a Pearson‑style correlation with the flattened ground‑truth mask $m^{(i)}\!\in\!\{0,1\}^{V}$ by cosine on zero‑mean vectors:
\[
r^{(i)}_{l,h}
=\Big\langle
\frac{a^{(i)}_{l,h}-\overline{a^{(i)}_{l,h}}}{\|a^{(i)}_{l,h}-\overline{a^{(i)}_{l,h}}\|_2},
\frac{m^{(i)}-\overline{m^{(i)}}}{\|m^{(i)}-\overline{m^{(i)}}\|_2}
\Big\rangle .
\]
Across $N$ samples this yields two heatmaps: the mean correlation $r_{l,h}=\tfrac{1}{N}\sum_i r^{(i)}_{l,h}$ and the \emph{frequency} $p_{l,h}=\tfrac{1}{N}\sum_i \mathbf{1}[\,r^{(i)}_{l,h}>\tau\,]$ that the per‑sample correlation exceeds a threshold $\tau$ (default $0.3$). We rank heads using a layer‑normalized score that rewards both strength and consistency, e.g.
\[
s_{l,h} \;=\; \mathrm{ReLU}\!\big(z_l(r_{l,h})\big)\cdot p_{l,h},
\]
where $z_l(\cdot)$ z‑scores within layer $l$ to remove layer‑scale effects. For models with grouped‑query attention (GQA), let $\mathcal{G}(l,k)$ denote the set of query heads tied to KV head $k$ in layer $l$; we collapse to KV granularity via
\[
s^{\mathrm{KV}}_{l,k} \;=\; \max_{h\in \mathcal{G}(l,k)} s_{l,h},
\]
and select the $28$ KV heads with the largest $s^{\mathrm{KV}}_{l,k}$ for subsequent use.

One visualization of the correlation-map–based selection is shown in Figure~\ref{fig:headselection_corrmap}. Compared to our proposed head-selection procedure (Figure~\ref{fig:headselection_ours_12}, under ratio 1:2), it tends to concentrate more heavily on a few middle-late layers.

\begin{figure*}[t]
    \centering
    \includegraphics[width=0.4\linewidth]{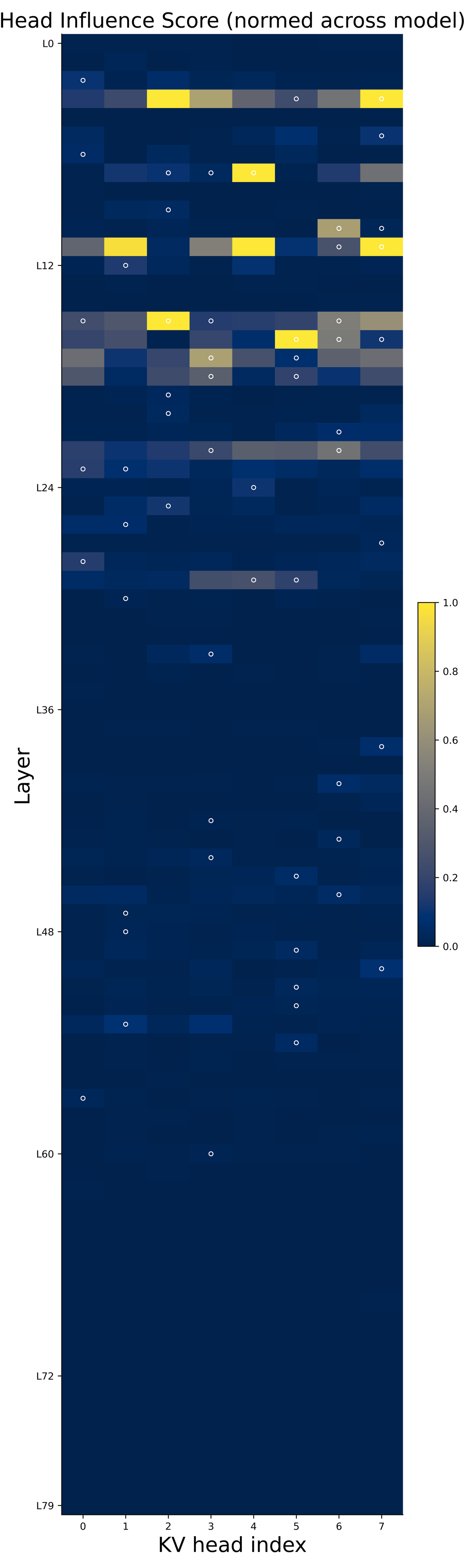}
    \vspace{-0.1in}
    \caption{\textbf{Head selection for Qwen2.5-VL-72B under a input-text:image token ratio of 1:2.}}
    \label{fig:headselection_72B}
\end{figure*}

\paragraph{More Head Selection Results \& Studies}
We list more visualizations for our head selection results, including one 72B model selection (Figure~\ref{fig:headselection_72B}), one LLaVA-Next-7B selection (Figure~\ref{fig:headselection_llava}), the head selections with different head budgets (Figure~\ref{fig:headselection_different_headbudgets}), and the head selections under different input-text:image token ratios (Figure~\ref{fig:headselection_7B_different_ratios}).

\input{figs/tab_appen_7B_params}

\input{figs/tab_appen_72B_params}

\clearpage



\input{figs/headselection_corrmap}

\input{figs/headselection_different_headbudget}
\input{figs/headselection_7B_different_ratios}

\clearpage

\section{Limitations and Discussions}
\label{append:limitations}

The proposed Image-LoRA introduces fundamental challenges for batching, as image-token and text-token lengths vary across samples and must be handled efficiently. Both training and inference further demand substantial kernel-level and hardware-level optimizations to achieve competitive throughput. In particular, because Image-LoRA applies updates only to visual-token spans, designing an efficient inference pathway becomes crucial: in its current form, Image-LoRA must dynamically locate visual-token spans for every inference sample. This behavior differs fundamentally from Std-LoRA, which merges weights before inference without requiring such runtime checks.

Image-LoRA is most naturally applicable to transformer-based VLMs that represent images as explicit visual-token sequences. For architectures such as Llama-3.2-Vision~\citep{meta2024llama}, which integrate visual information via cross-attention directly into the text representation, Image-LoRA behaves much more like Std-LoRA, since the notion of a distinct visual-token span becomes less well-defined.

Regarding experimental scale, our current results are limited to on the order of a thousand samples due to constrained compute and the need to validate multiple components. A more comprehensive evaluation would ideally follow the large-scale setting of \citet{schulman2025lora}, which employs extensive data and dense sweeps over experimental settings, such as LoRA-rank variations.

Moreover, a detailed analysis of how Image-LoRA reshapes attention patterns across heads and layers—and how these patterns differ from those induced by Std-LoRA—could clarify its underlying mechanisms and inform further improvements. Beyond that, an open question is whether the dynamics of these pattern changes during training differ between Image-LoRA and Std-LoRA.

Finally, understanding the interaction between Image-LoRA and head selection is an important avenue for future work. Our head-selection procedure is central to reducing the number of trainable parameters and, consequently, adapter-only FLOPs. It would be valuable to study how the selected heads evolve over the course of Image-LoRA training, particularly in web-scale settings where different subsets of attention heads may be optimal at different training stages. Moreover, this paper does not systematically examine how different head-selection methods affect performance: we focus on a single, simple head-selection scheme, and alternative criteria may lead to substantially different accuracy–efficiency trade-offs.

\section{Broader Impact}
\label{appen:broader-impact}
This paper proposes Image-LoRA, a parameter-efficient fine-tuning method for vision-language models that adapts only the value path within the visual-token span and a selected subset of attention heads, reducing adapter-only training FLOPs while maintaining performance on screen-centric grounding/referring benchmarks and preserving text-only reasoning (e.g., GSM8K). By lowering the compute cost of adapting VLMs, this approach may reduce energy use and broaden access to customization and deployment in resource-constrained settings for VLMs.

%% file: sec/method_appendix.tex
\newcommand{\hkv}{\ensuremath{H}}
\newcommand{\nchosen}{\ensuremath{N_{\text{chosen}}}}
\newcommand{\vlj}{\ensuremath{V^{(h)}}}
\newcommand{\variance}{\ensuremath{\sigma^2}}

\section{Method Details}

\subsection{Head Selection Procedure (details)}
\label{app:head-selection-details}

\paragraph{Visual-only gradients \& Different ratios} For our head selection procedure, gradients are restricted to the visual token slice when estimating the head importance score $I(h)$ and head diversity feature $F(h)$ for each head $h$. For the same model but different input text:image token ratios, we run the head selection procedure once for each ratio.

\paragraph{Layerwise budgeting.}
Using the per‑head importance $I(h)$ defined in the main text, we first accumulate per‑layer mass
$\Phi_L = \sum_{h\in L} I(h)$ and form normalized allocation weights
$p_L = \frac{\Phi_L^\tau}{\sum_{L'} \Phi_{L'}^\tau}$ with temperature $\tau\!\ge\!0$, so that $p_L$ encodes the fraction of the global head budget assigned to layer $L$ (and $\sum_L p_L = 1$).
Briefly, $\tau=0$ yields uniform-by-layer allocation; $\tau=1$ is proportional to mass (linear);
and $\tau\in(0,1)$ compresses differences in layer mass (still favoring high-mass layers, but more gently).
In our experiments as shown in Appendix~\ref{append:exp:main}, we find $\tau=0.5$ to be a balanced choice.
Given a global budget of $K_{\text{sel}}$ heads (like $K_{\text{sel}}=28$ we used in the main experiment), these allocation weights $p_L$ define target per-layer quotas. We first compute initial rounded quotas
$\tilde{k}_L = \left\lfloor K_{\text{sel}}\, p_L + \tfrac{1}{2}\right\rfloor$,
cap each by capacity $\hkv$,
\[
k_L \;\leftarrow\; \min\!\big(\tilde{k}_L,\; \hkv\big),
\]
then greedily adjust $\{k_L\}$ so that $\sum_L k_L = K_{\text{sel}}$ exactly:
if $\sum_L k_L<K_{\text{sel}}$ we add heads to layers in descending order of $\Phi_L$ while respecting capacities; if $\sum_L k_L>K_{\text{sel}}$ we remove heads from layers in ascending order of $\Phi_L$ until the sum matches.
(When $\sum_L \Phi_L{=}0$, we fall back to capacity‑proportional $p_L$.) The resulting $\{k_L\}$ are the per‑layer budgets used by the within‑layer candidate selection step described next.

\paragraph{Candidate pool and diversity‑aware selection.}
Within layer $L$, given its budget $k_L$, we first assemble a candidate pool
\[
k_L' \;=\; \min\!\big(\hkv,\,\lceil \rho\,k_L\rceil\big),\qquad
P_L \;=\; \operatorname*{Top}_{k_L'} \{ I(h): h\!\in\!L\},
\]
where $\rho\!\ge\!1$ is an expansion factor and $I(h)$ is the importance score from the main text. 
As $\rho \to 1^+$, selection becomes primarily importance‑driven;  at $\rho=1$ it reduces to pure top‑$k$ by importance within layer $L$. We set $\rho=2$ for our main experiments and studied it in our ablation experiments. To discourage near‑duplicates, we use the head diversity feature $F(h) = (\nabla_{B^{(h)}} \ell) \odot (\nabla_{B^{(h)}} \ell)$ (also defined in Section~\ref{sec:method-head-selection}) and perform \emph{cosine farthest‑first} selection on row‑normalized $\{F(h): h\in P_L\}$:
initialize with the highest‑importance head from $P_L$, then iteratively add the candidate that maximizes its minimum cosine distance to the already‑selected set.
This returns a diverse subset $\mathcal{H}^{(L)}_{\text{chosen}}$ of size $k_L$. We show the importance of including the diversity term in the experimental results presented in the Appendix~\ref{append:exp:main}. 

\begin{algorithm}[h]
  \caption{Head selection via Head Influence Scores Estimation}
  \label{alg:ohi}
  \begin{algorithmic}[1]
    \Require probe set \(\mathcal{D}\), global KV budget \(K_{\text{sel}}\), temperature \(\tau\), diversity multiplier \(\rho\)
    \For{each layer \(L\)}
      \State rank‑1 approximation: shared \(A^{(L)}\!\in\!\mathbb{R}^{\dhid\times 1}\), per‑KV \(B^{(L)}_{j}\!\in\!\mathbb{R}^{1\times \dhead}\);
             freeze base, set \(\gamma^{(L)}\!\gets\!1\); restrict gradients to the visual token slice.
      \State initialize importance \(I^{(L)}_{j}\!\gets\!0\) and diversity feature \(F^{(L)}_{j}\!\gets\!0\) for \(j{=}1,\dots,H^{(L)}_{kv}\).
    \EndFor
    \For{each \((X,y)\in\mathcal{D}\)}
      \State forward with CE loss; one backward pass
      \For{each layer \(L\), head \(j\)}
        \State \(\Delta\gets \nabla_{B^{(L)}_{j}}\ell\in\mathbb{R}^{\dhead}\)
        \State \(I^{(L)}_{j}\pluseq \|\Delta\|_2^2\);\quad \(F^{(L)}_{j}\pluseq \Delta^{\odot 2}\)
      \EndFor
    \EndFor
    \State compute \(\Phi_L=\sum_j I^{(L)}_j\) and \(p_L=\Phi_L^\tau/\sum_{L'} \Phi_{L'}^\tau\)
           \Comment{if \(\sum_{L'} \Phi_{L'}{=}0\), set \(p_L \propto H^{(L)}_{kv}\)}
    \State \(\tilde{k}_L\gets \left\lfloor K_{\text{sel}}\,p_L + \tfrac{1}{2}\right\rfloor\);\quad \(k_L\gets \min(\tilde{k}_L,\,\hkv)\)
    \State greedily adjust \(\{k_L\}\) so that \(\sum_L k_L = K_{\text{sel}}\) (add by descending \(\Phi_L\); remove by ascending \(\Phi_L\))
    \For{each layer \(L\)}
      \State form pool \(P_L\): top \(\min(\hkv,\lceil \rho\,k_L\rceil)\) by importance \(I^{(L)}_j\); row‑normalize diversity features \(F^{(L)}_j\gets F^{(L)}_j/\|F^{(L)}_j\|_2\)
      \State select \(k_L\) heads from \(P_L\) via cosine farthest‑first (initialized with the highest‑importance head) on \(\{F^{(L)}_j\}\), yielding \(\mathcal{H}^{(L)}_{\text{chosen}}\)
      \State map KV\(\rightarrow\)Q (if grouped‑query is used): \(\mathcal{H}^{(L)}_q \gets \bigcup_{j\in\mathcal{H}^{(L)}_{\text{chosen}}}\Gamma^{(L)}(j)\)
    \EndFor
    \State \Return \(\{\mathcal{H}^{(L)}_{q}\}_L\) \Comment{selected Q‑head sets used by Image‑LoRA training}
  \end{algorithmic}
\end{algorithm}

\paragraph{Complexity.}
The procedure requires one forward+backward pass per probe example with gradients confined to the visual slice. Storage per layer is $O(\hkv\,\dhead)$ for diversity features and $O(\hkv)$ for scalar scores. Farthest‑first selection over the pool is $O(k'_L\,\dhead)$ for a single sweep.

\subsection{Head-selection Normalization.}
\label{sec:layer_normalization}
Standard LoRA implicitly activates \emph{all} KV heads per layer, i.e., \(\nchosen\!=\!\hkv\), so the per‑layer update magnitude is implicitly comparable across layers.
With head selection, \(\nchosen\) varies; to prevent the summed increment from growing with \(\nchosen\), we scale the update as
\[
s \;=\; \tfrac{\alpha}{r}\,\gamma\,\nu\!\big(\nchosen\big).
\]
A variance-preserving choice follows from a simple energy argument.
Let \(\Delta \vlj\) be the per-head increment (for head \(j\)) and suppose \(\mathbb{E}[\Delta \vlj]=0\) with \(\mathbb{E}\!\left[\|\Delta \vlj\|_F^2\right]=\sigma_L^2\), and approximate different heads as uncorrelated.
Then
\[
\mathbb{E}\!\left[\Big\|\sum_{h\in \mathcal{H}^{(L)}_{\text{chosen}}} \Delta \vlj\Big\|_F^2\right]
\;\;=\; \nchosen\,\variance.
\]
After scaling by \(\nu(\nchosen)\), the expected energy becomes \(\nu(\nchosen)^2\,\nchosen\,\variance\).
Setting this equal to \(\variance\) yields the variance-preserving rule
\[
\nu\!\big(\nchosen\big) \;=\; \frac{1}{\sqrt{\nchosen}}.
\]
If heads exhibit positive average pairwise correlation \(\rho\) (so \(\mathbb{E}\langle \Delta V^{(i)},\Delta V^{(j)}\rangle \approx \rho\variance\) for \(i\!\neq\!j\)), the same calculation gives
\[
\mathbb{E}\!\left[\Big\|\sum_{h\in \mathcal{H}^{(L)}_{\text{chosen}}} \Delta \vlj\Big\|_F^2\right]
 \approx 
\big(\nchosen + \rho\,\nchosen(\nchosen - 1)\big)\variance
\]
and an energy-matching scale
\[
\nu\!\big(\nchosen\big) = \big(\nchosen + \rho\,\nchosen(\nchosen - 1)\big)^{-1/2}
\]
When \(\rho\) is unknown, the \(\tfrac{1}{\sqrt{\nchosen}}\) rule is a robust approximation; the learned gate \(\gamma\) absorbs residual mismatch.

\subsection{Visual-token-only finetuning with unchanged attention.}
\label{sec:normalization_equivalence}
Let the self-attention in layer \(L\) project inputs \(X\!\in\!\mathbb{R}^{T \times \dhid}\) to
\(V \!\in\!\mathbb{R}^{T \times H\times \dhead}\) via \(v_{\rm proj}:\mathbb{R}^{\dhid}\!\to\!\mathbb{R}^{H\dhead}\).
We adapt only the \emph{value (V)} path on the visual‑token index set \(\mathcal{I}_v\).
The attention probabilities
\(a^{(h)}_{i,j}=\operatorname{softmax}_j(\langle q^{(h)}_i,k^{(h)}_j\rangle/\sqrt{\dhead})\)
remain unchanged, while the output becomes
\[
\tilde{o}^{(h)}_i=\sum\nolimits_j a^{(h)}_{i,j}\,\tilde{v}^{(h)}_j,
\qquad
\tilde{v}^{(h)}_j = v^{(h)}_j + \Delta v^{(h)}_j.
\]
Head‑wise scaling of values by \(c_j\) satisfies
\[
\sum\nolimits_j a^{(h)}_{i,j}(c_j v^{(h)}_j)
\;=\;
\sum\nolimits_j (c_j a^{(h)}_{i,j})\,v^{(h)}_j,
\]
so modifying values is equivalent to multiplying attention weights without re-normalizing the softmax. In our ablation studies, we applied the modification to \(K\) before normalization —thus changing the probability distribution—and found that this consistently degraded performance. Restricting updates to \(\mathcal{I}_v\) also reduces training FLOPs roughly by \(T_v/T\), which is substantial for text-heavy prompts.

\subsection{Grouped Query Attention (GQA)}
\label{sec:GQA}
For a transformer layer \(L\) with hidden width \(\dhid\), head dimension \(\dhead\), \(H_q\) query heads and \(H_{kv}\) key/value heads, we assume grouped‑query attention with grouping factor \(g\), so that \(H_q = g\,H_{kv}\).
We denote the \emph{selected} query heads by
\(\mathcal{H}_q \subset \{0,\dots,H_q\!-\!1\}\),
and their induced chosen KV heads by
\[
\mathcal{H}_{chosen} \;=\; \big\{ \lfloor h/g \rfloor : h \in \mathcal{H}_q \big\},
\qquad
\nchosen \triangleq \big|\mathcal{H}_{chosen}\big|.
\]
Thus, multiple query heads may share the same KV head; \(\nchosen\) counts the number of \emph{unique} KV heads.
We denote the visual‑token index set by \(\mathcal{I}_v\) and its length by \(T_v = |\mathcal{I}_v|\).
Throughout, we use the KV index purely to address heads; we adapt \(V\) only in main experiments, while \(K\) is included only for ablation studies.

\subsection{Parameter Counts on \textsc{Qwen2.5-VL-7B} (GQA).}
\label{sec:parameter_counts}
\textsc{Qwen2.5-VL-7B} uses grouped-query attention: \(H_q=28\) query heads share \(H_{kv}=4\) key/value heads (we use the KV index to address heads; only \(V\) is adapted).
With \(\dhid{=}3584\), \(\dhead{=}128\), \(N_L{=}28\) layers, rank \(r{=}8\), and an illustrative selection \(\nchosen = 4\) KV heads per layer:
\[
\begin{aligned}
\#A &= \dhid\,\times r = 3584 \times 8 = 28{,}672,\\
\#B^{(h)} &= \nchosen\, \times r\, \times \dhead = 4 \times 8 \times 128 = 4{,}096.
\end{aligned}
\]
Including one scalar \(\gamma\) gives \(28{,}672 + 4{,}096 + 1 = 32{,}769\) parameters per layer and \(32{,}769 \times N_L = 917{,}532\) across the stack. The dominant \(\dhid \times r\) term motivates sharing \(A\); head-specific \(B\) ($r\, \times \dhead$) preserves per-head expressivity at a cost proportional to \(N^{\text{chosen}}\).

%% file: figs/tab_appen_headselection.tex
\begin{table}[!t]
  \centering
  \setlength{\tabcolsep}{4pt}
  \caption{\textbf{Ablations on layerwise budgeting $\tau$ and diversity factor $\rho$ on Qwen2.5-VL-7B.}
Columns report accuracy (\%) at different text:image
  token ratios. Defaults underlined \underline{($\tau=0.5, \rho=2.0$)}. }
  \label{tab:abl-tau-diversity}
  \begin{tabular}{l c c c c}
    \toprule
    Method & 1:5 & 1:4 & 1:3 & 1:2 \\
    \midrule
    \multicolumn{5}{l}{\emph{Layerwise budgeting (diversity-aware, $\rho = 2.0$ fixed)}} \\
    \midrule
    $\tau = 0$ \ (uniform-by-layer)      & 34.77 & 27.88 & 23.06 &  16.87 \\
    \underline{$\tau = 0.5$} \             & \textbf{37.35} & \textbf{30.98} & \textbf{24.44} & 18.24 \\
    $\tau = 1$ \ (mass-proportional)    & 35.11 & 28.40 & 23.92 & \textbf{18.59} \\
    \midrule
    \multicolumn{5}{l}{\emph{Diversity vs.\ importance-only ($\tau = 0.5$ fixed)}} \\
    \midrule
    Importance-only ($\rho = 1$)        & 35.11 & 28.40 & 23.75 & 17.9 \\
    \underline{Diversity-aware ($\rho = 2.0$)} & \textbf{37.35} & \textbf{30.98} & \textbf{24.44} & \textbf{18.24} \\
    Diversity-aware ($\rho = 3.0$)      & 34.08 & 29.09 & 22.55 & 16.52 \\
    \bottomrule
  \end{tabular}
\end{table}

%% file: figs/headselection_different_tau.tex
\begin{figure*}[ht]
    \centering
    
    \begin{subfigure}[b]{0.32\linewidth}
        \centering
        \includegraphics[width=\linewidth]{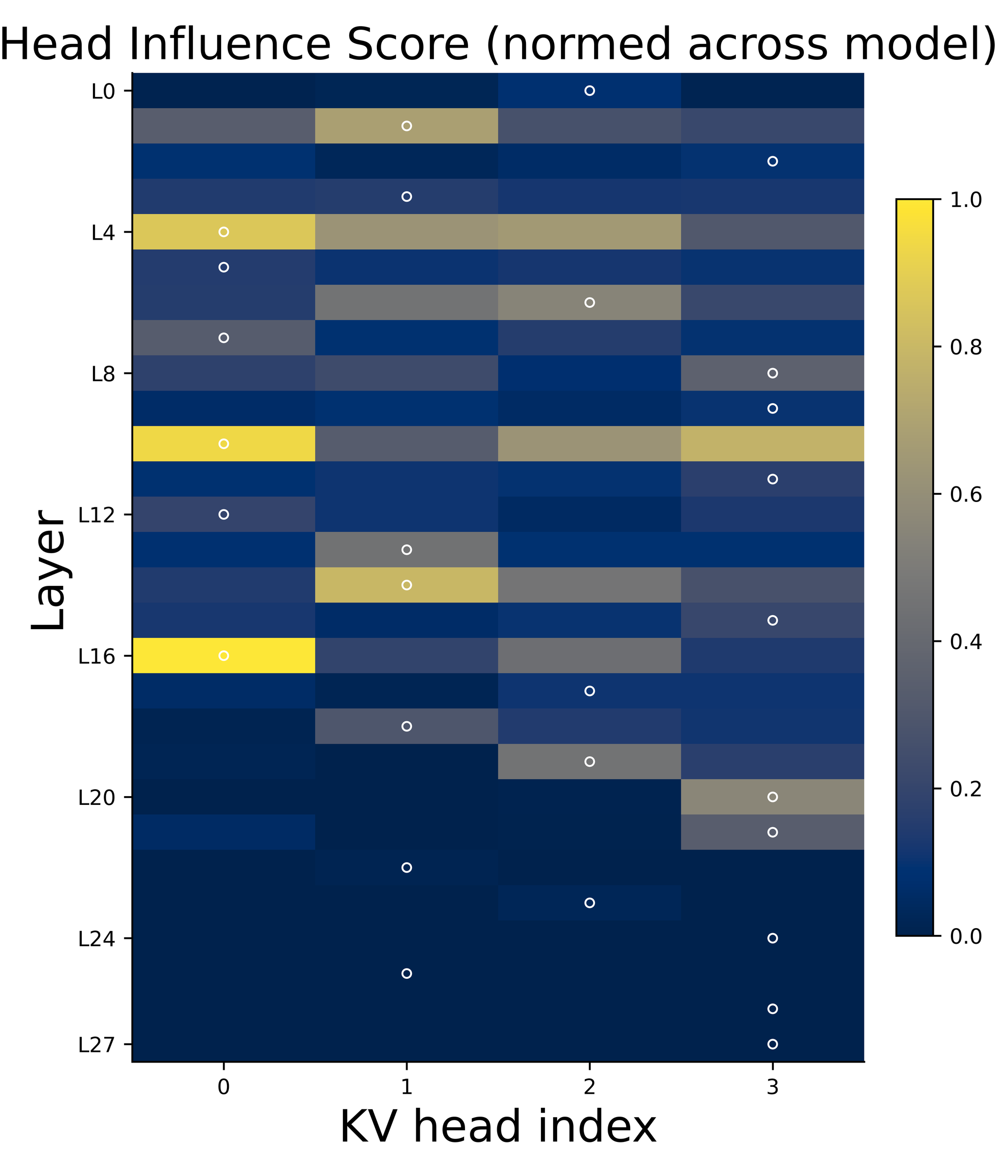}
        \caption{Head selection with $\tau = 0$.}
        \label{fig:ohi_scores_1head}
    \end{subfigure}
    \hfill
    \begin{subfigure}[b]{0.32\linewidth}
        \centering
        \includegraphics[width=\linewidth]{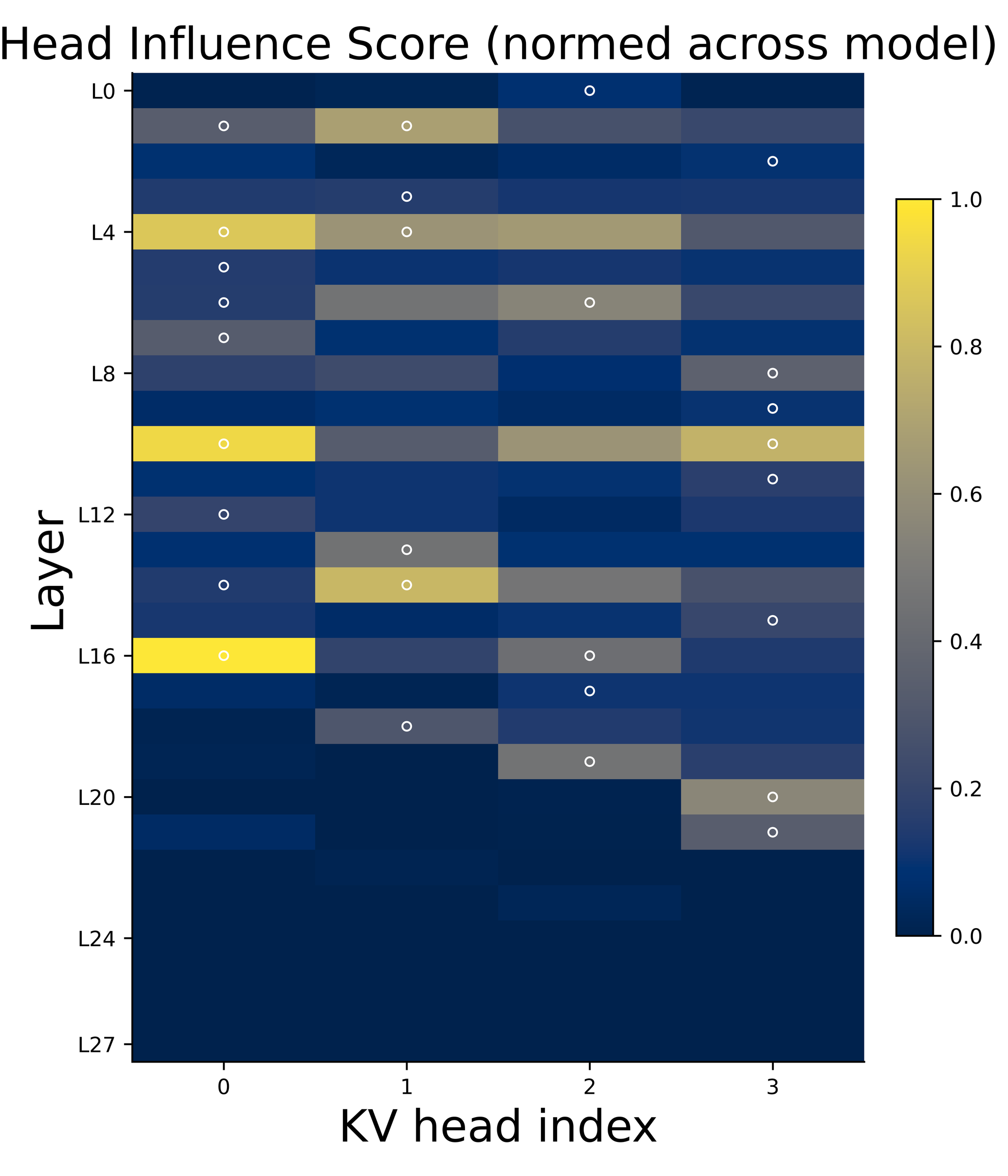}
        \caption{Head selection with $\tau = 0.5$ (default).}
        \label{fig:ohi_scores_7heads}
    \end{subfigure}
    \hfill
    \begin{subfigure}[b]{0.32\linewidth}
        \centering
        \includegraphics[width=\linewidth]{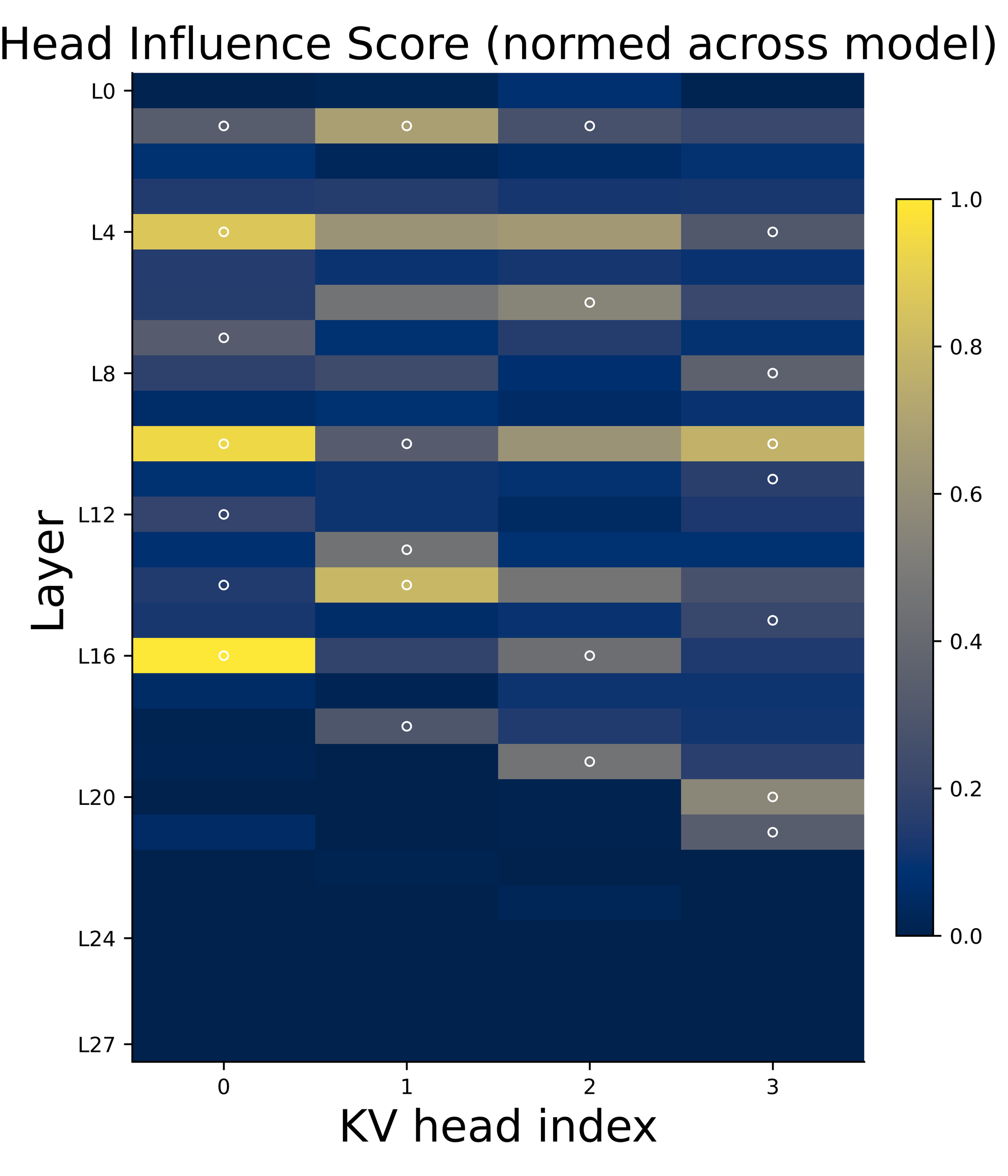}
        \caption{Head selection with $\tau = 1$.}
        \label{fig:ohi_scores_28heads}
    \end{subfigure}
    \vspace{-0.1in}
    \caption{\textbf{Head selection patterns with different $\tau$.} The head selection procedure uses default hyper-parameters $\rho = 2$. All results are obtained on ScreenSpot-Pro using the 1:2 input-text:image token ratio.
    Intuitively, $\tau = 0$ yields an \emph{approximately uniform} allocation across layers, and $\tau = 1$ makes the budget \emph{proportional to mass}~$\Phi_L$,
We use $\tau = 0.5$ in our main experiments. For Qwen2.5-VL-7B with $28$ layers and $K_{\text{sel}} = 28$, the $\tau = 0$ setting reduces to select exactly one KV head per layer — the one with the highest importance score. 
    }
    \label{fig:headselection_different_headbudgets}
\end{figure*}

%% file: figs/headselection_different_rho.tex
\begin{figure*}[ht]
    \centering
    
    \begin{subfigure}[b]{0.32\linewidth}
        \centering
        \includegraphics[width=\linewidth]{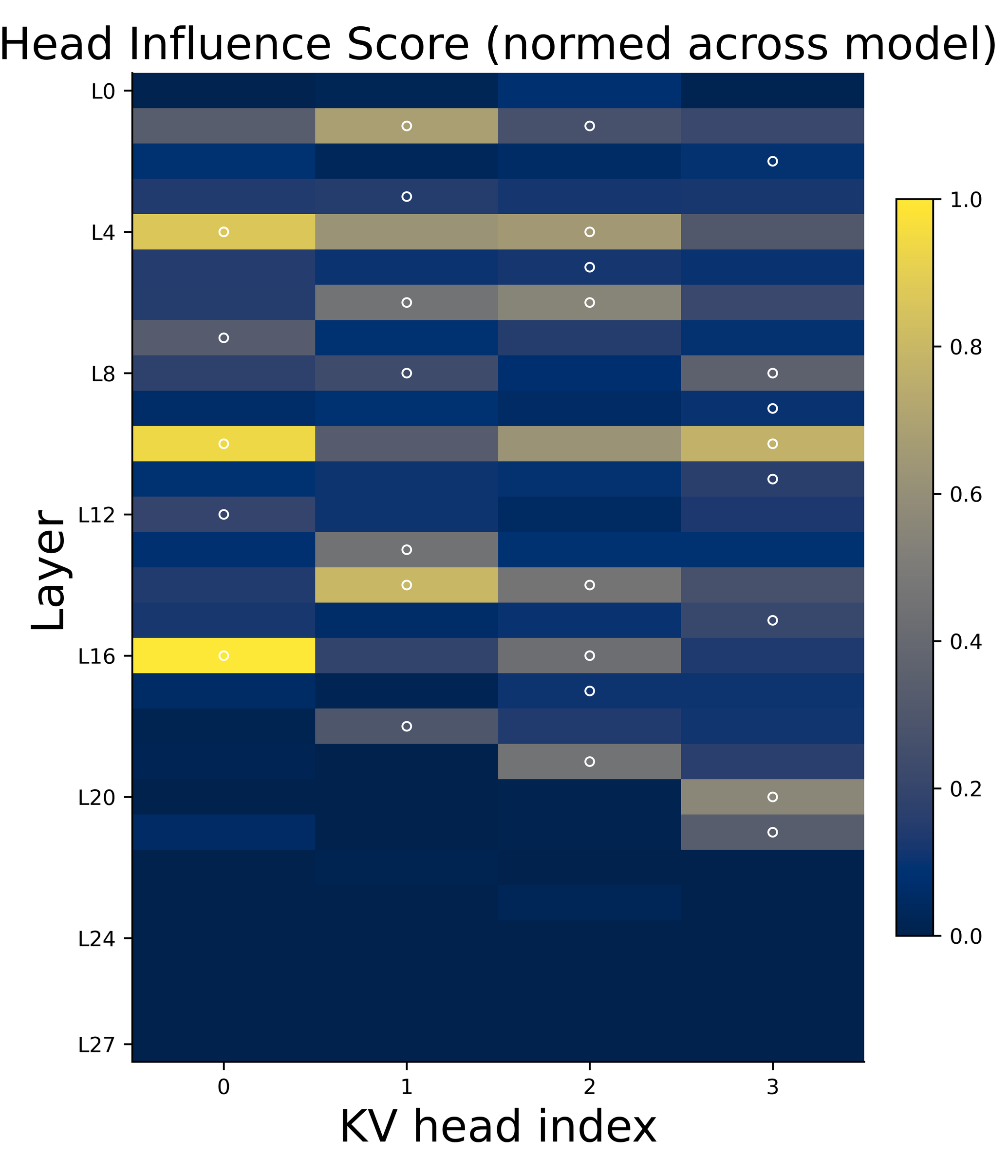}
        \caption{Head selection with $\rho = 1$.}
    \end{subfigure}
    \hfill
    \begin{subfigure}[b]{0.32\linewidth}
        \centering
        \includegraphics[width=\linewidth]{figs/ohi_scores_globalnorm_12.png}
        \caption{Head selection with $\rho = 2$ (default).}
    \end{subfigure}
    \hfill
    \begin{subfigure}[b]{0.32\linewidth}
        \centering
        \includegraphics[width=\linewidth]{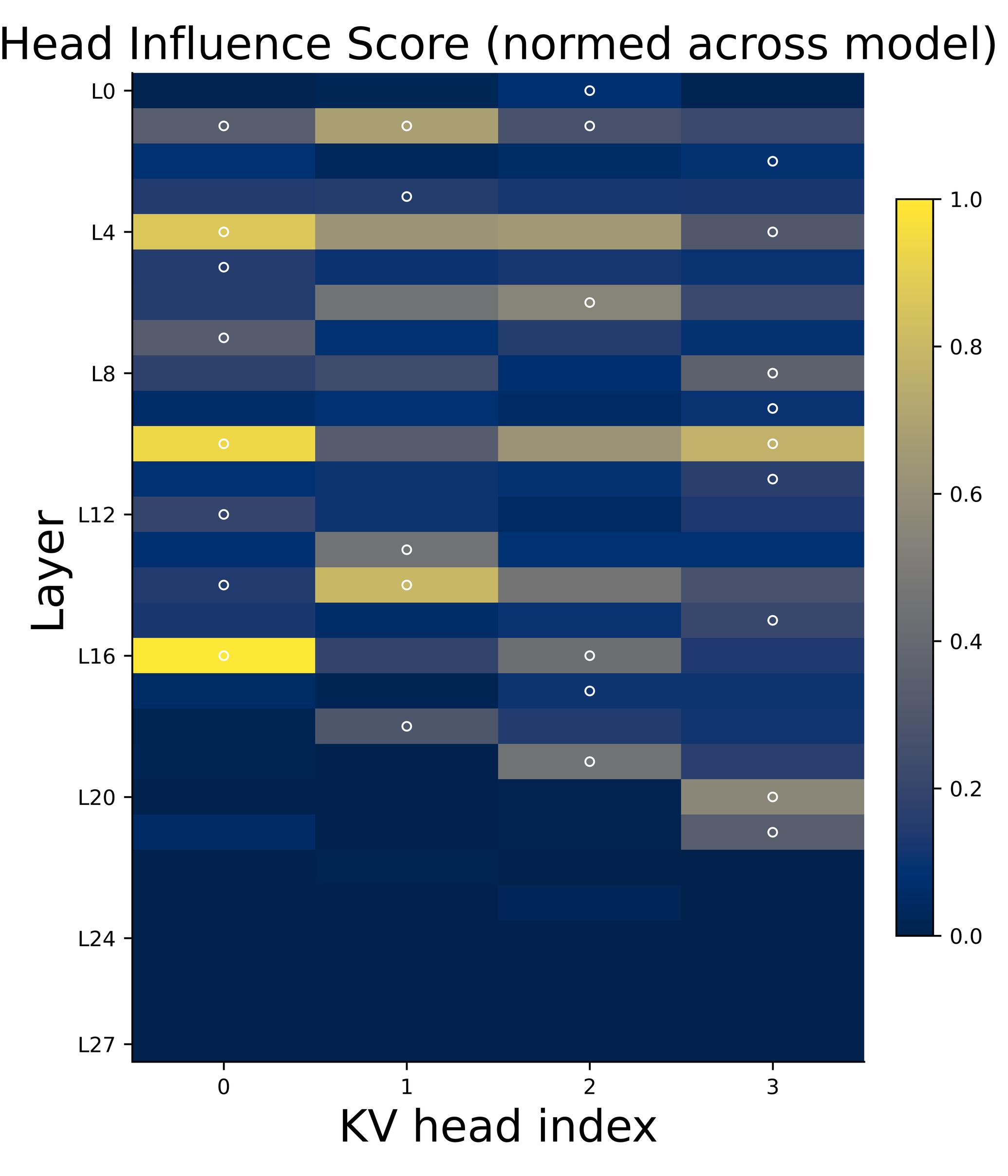}
        \caption{Head selection with $\rho = 3$.}
    \end{subfigure}
    \vspace{-0.1in}
    \caption{\textbf{Head selection patterns with different $\rho$.} The head selection procedure uses default hyper-parameters $\tau = 0.5$. All results are obtained on ScreenSpot-Pro using the 1:2 input-text:image token ratio. We use $\rho = 2$ in our main experiments. $\rho = 1$ reduces exactly to pure top-$k_L$ by importance, while larger $\rho$ allows more diversity at a small cost in $I(h)$.}
    \label{fig:headselection_different_rho}
\end{figure*}

%% file: figs/tab_appen_7B_params.tex
\begin{table*}[t]
\centering
\caption{\textbf{Trainable parameters (M) and per-example FLOPs (G) - Qwen2.5-VL-7B}. We test on ScreenSpot-Pro across input-text:image token ratios for the Qwen2.5-VL-7B model. Image-LoRA is V-only configurations.}
\setlength{\tabcolsep}{8pt}
\resizebox{\textwidth}{!}{%
\begin{tabular}{l*{4}{cc}}
\hline
Text:Img Token Ratio & \multicolumn{2}{c}{1:5} & \multicolumn{2}{c}{1:4} & \multicolumn{2}{c}{1:3} & \multicolumn{2}{c}{1:2} \\
Avg \#Image Tokens & \multicolumn{2}{c}{4195} & \multicolumn{2}{c}{3367} & \multicolumn{2}{c}{2534} & \multicolumn{2}{c}{1677} \\
\cline{2-3}\cline{4-5}\cline{6-7}\cline{8-9}

& Params & FLOPs 
& Params & FLOPs 
& Params & FLOPs 
& Params & FLOPs  \\
\textbf{Method}
& (M) & (G) & (M) & (G) & (M) & (G) & (M) & (G) \\
\hline
Image-LoRA  & 0.6298 & 15.85 & 0.6298 & 12.72 & 0.6298 &  9.576 & 0.6298 &  6.335 \\
Std-LoRA V-only     & 0.9175 & 27.83 & 0.9175 & 23.27 & 0.9175 & 18.68  & 0.9175 & 13.96 \\
Std-LoRA QV         & 2.523  & 76.53 & 2.523  & 63.99 & 2.523  & 51.38  & 2.523  & 38.38 \\
Std-LoRA QKVO       & 5.046  & 153.1 & 5.046  & 128.0 & 5.046  & 102.8  & 5.046  & 76.76 \\
\hline
\end{tabular}
}
\label{tab:sreenspot7b_lora_param_flops}
\end{table*}

%% file: figs/tab_appen_72B_params.tex
\begin{table*}[t]
\centering
\caption{\textbf{Trainable parameters (M) and per-example FLOPs (G) - Qwen2.5-VL-72B}. We test on ScreenSpot-Pro across input-text:image token ratios for the Qwen2.5-VL-72B model. Image-LoRA is V-only configurations.}
\setlength{\tabcolsep}{8pt}
\resizebox{\textwidth}{!}{%
\begin{tabular}{l*{4}{cc}}
\hline
Text:Img Token Ratio & \multicolumn{2}{c}{1:5} & \multicolumn{2}{c}{1:4} & \multicolumn{2}{c}{1:3} & \multicolumn{2}{c}{1:2} \\
Avg \#Image Tokens   & \multicolumn{2}{c}{4195} & \multicolumn{2}{c}{3367} & \multicolumn{2}{c}{2534} & \multicolumn{2}{c}{1677} \\
\cline{2-3}\cline{4-5}\cline{6-7}\cline{8-9}
& Params & FLOPs 
& Params & FLOPs 
& Params & FLOPs 
& Params & FLOPs  \\
\textbf{Method}
& (M) & (G) & (M) & (G) & (M) & (G) & (M) & (G) \\
\hline
Image-LoRA        & 3.012 & 75.80 & 2.945 & 59.49 & 2.881 & 43.80 & 2.877 & 28.94 \\
Std-LoRA V-only   & 5.898 & 178.9 & 5.898 & 149.6 & 5.898 & 120.1 & 5.898 & 89.72 \\
Std-LoRA QV       & 16.38 & 496.9 & 16.38 & 415.5 & 16.38 & 333.7 & 16.38 & 249.2 \\
Std-LoRA QKVO     & 32.77 & 993.9 & 32.77 & 831.0 & 32.77 & 667.3 & 32.77 & 498.4 \\
\hline
\end{tabular}
}
\label{tab:sreenspot72b_lora_param_flops}
\end{table*}

%% file: figs/headselection_corrmap.tex
\begin{figure*}[h]
    \centering
    \begin{minipage}[t]{0.64\textwidth}
        \centering
        \begin{subfigure}[t]{0.48\textwidth}
            \centering
            \includegraphics[width=\linewidth]{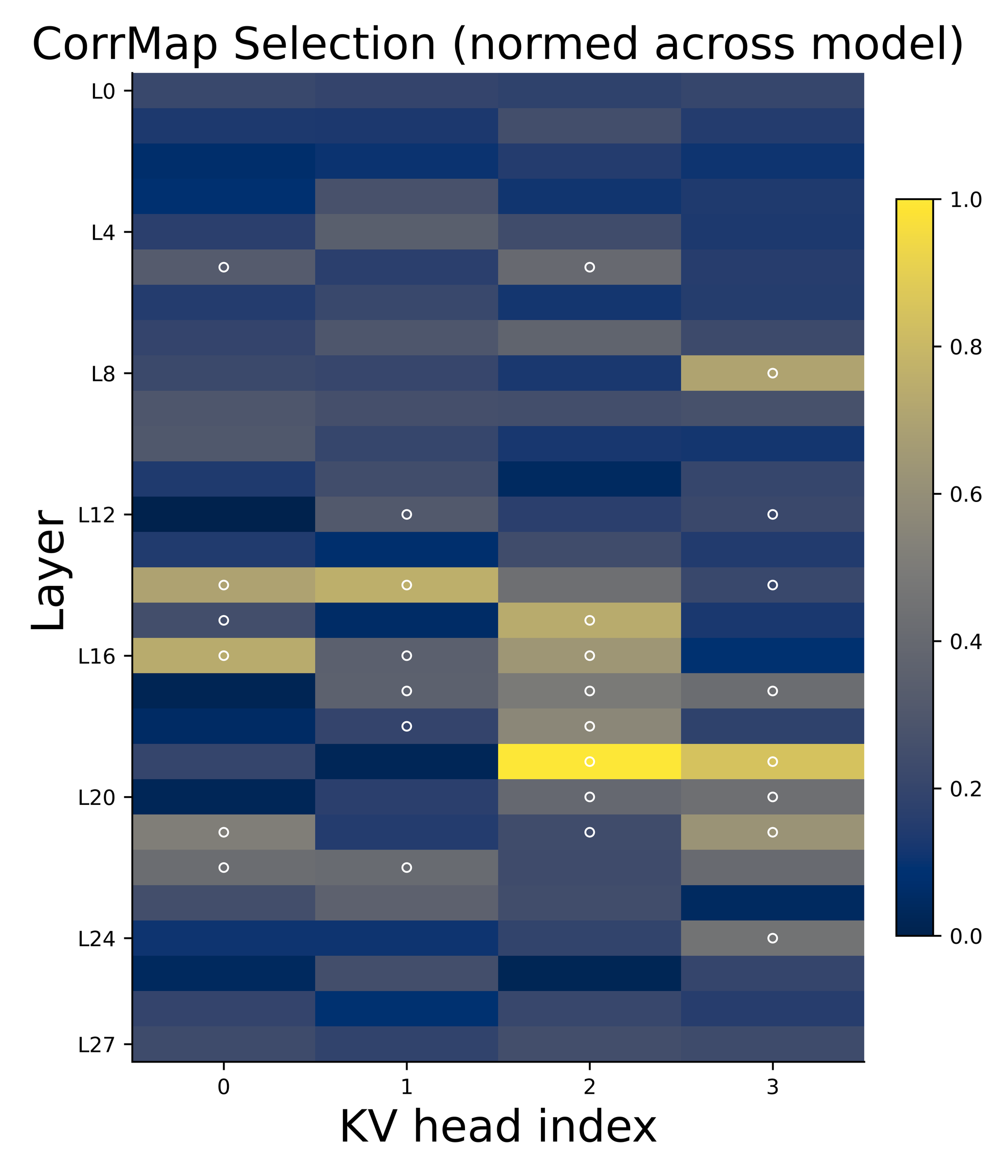}
            \caption{\textbf{Correlation-based head selection.}}
            \label{fig:headselection_corrmap}
        \end{subfigure}
        \begin{subfigure}[t]{0.48\linewidth}
            \centering
            \includegraphics[width=\linewidth]{figs/ohi_scores_globalnorm_12.png}
            \caption{\textbf{Our head selection.}}
            \label{fig:headselection_ours_12}
        \end{subfigure}
        \caption{\textbf{Comparison between the correlation-map–based approach and our proposed head-selection method.}} 
    \end{minipage}
    \hfill
    \begin{minipage}[t]{0.32\textwidth}
        \centering
        \includegraphics[width=0.9\linewidth]{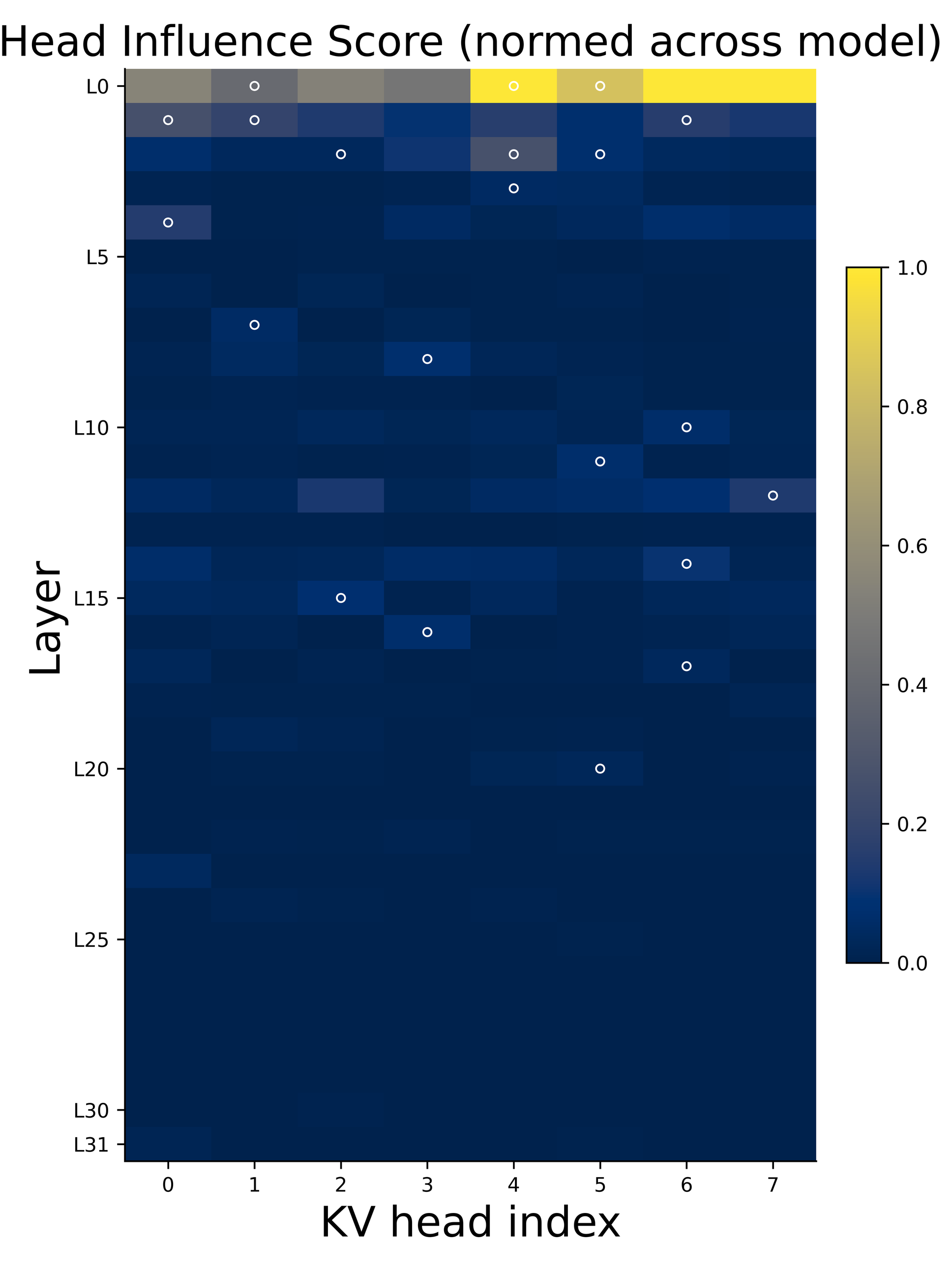}
        \caption{\textbf{Head selection for LLaVa-Next-7B under 1:2 text:image ratio.}}
        \label{fig:headselection_llava}
    \end{minipage}
\end{figure*}

%% file: figs/headselection_different_headbudget.tex
\begin{figure*}[h]
    \centering
    
    \begin{subfigure}[b]{0.32\linewidth}
        \centering
        \includegraphics[width=\linewidth]{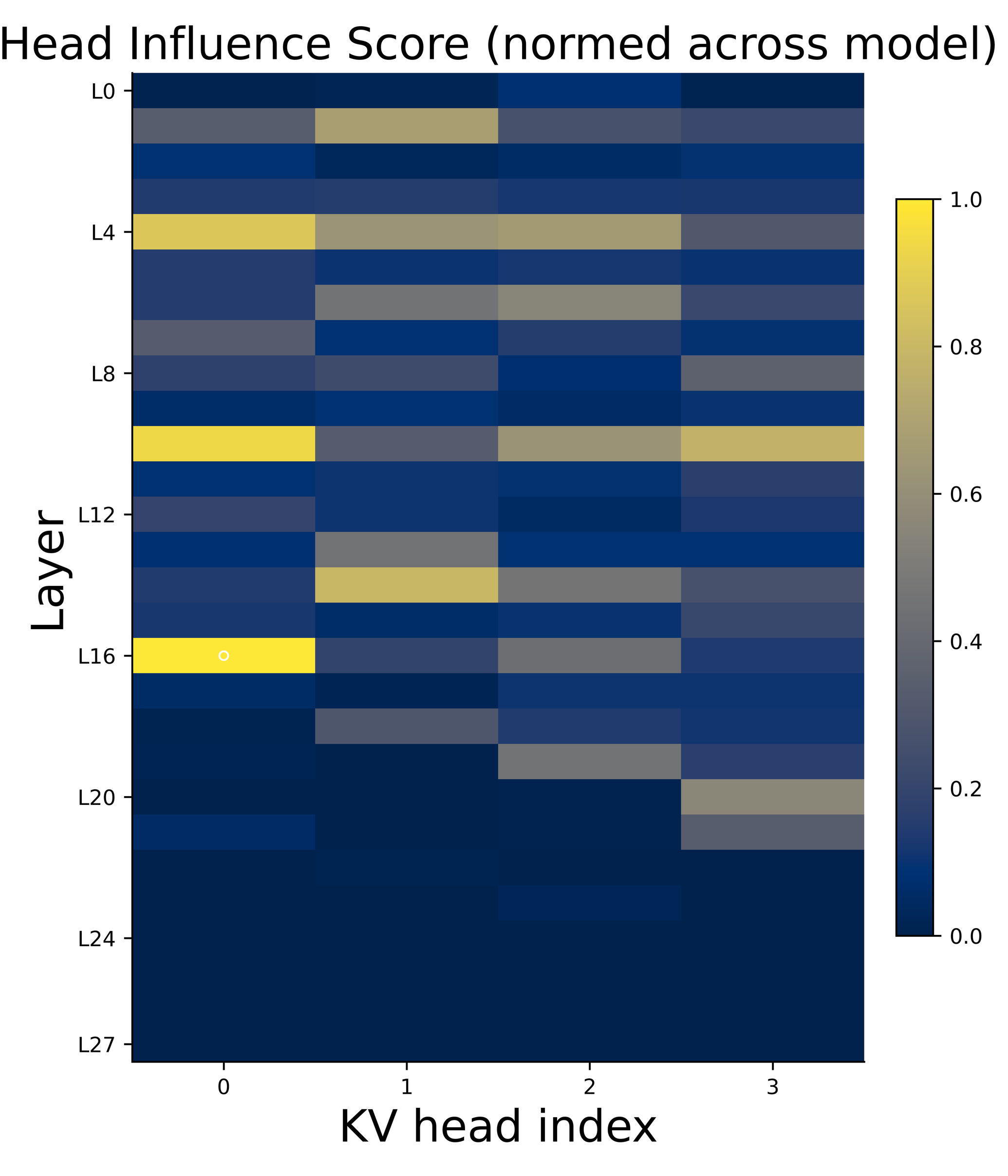}
        \caption{1 selected head $K_{sel}=1$.}
    \end{subfigure}
    \hfill
    \begin{subfigure}[b]{0.32\linewidth}
        \centering
        \includegraphics[width=\linewidth]{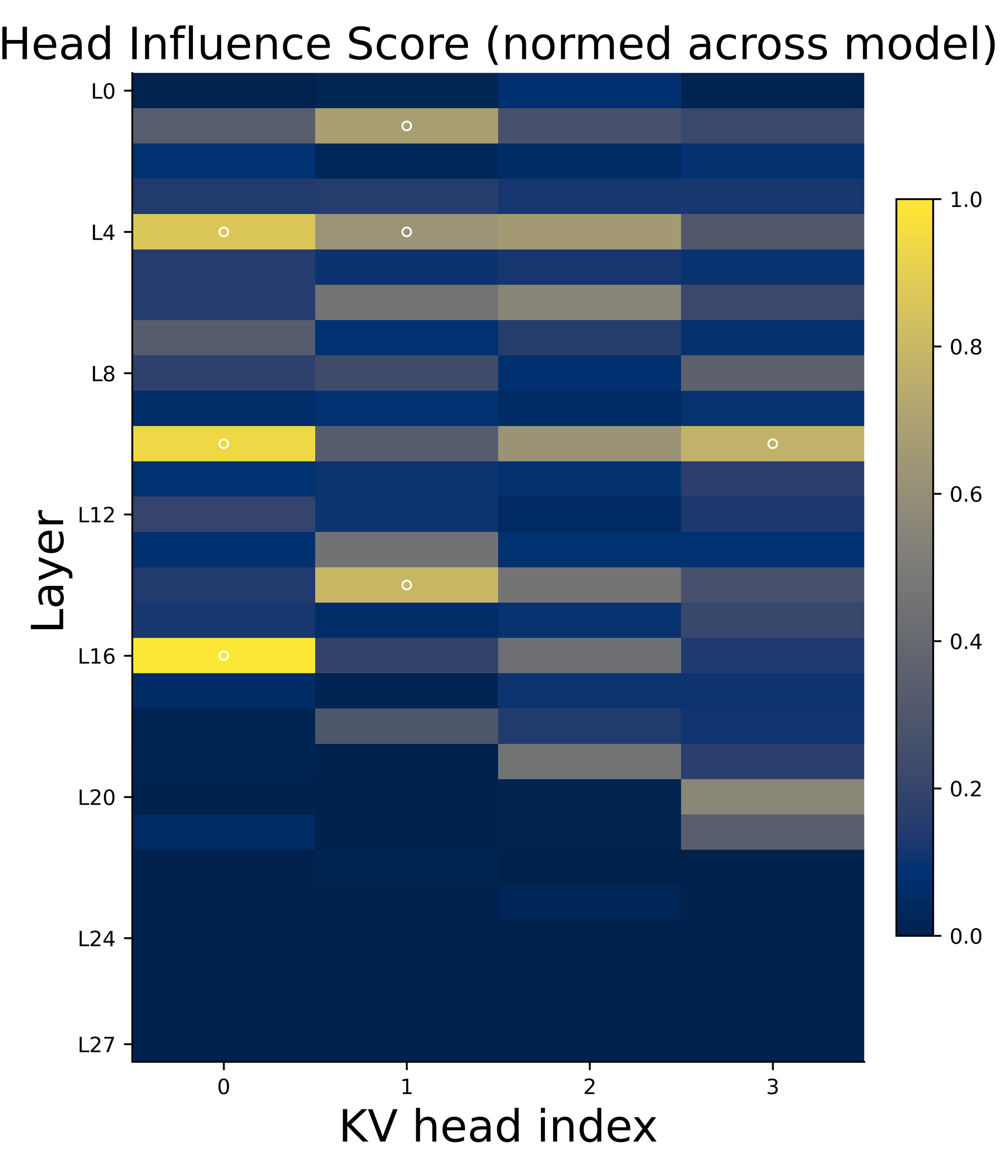}
        \caption{7 selected heads $K_{sel}=7$.}
    \end{subfigure}
    \hfill
    \begin{subfigure}[b]{0.32\linewidth}
        \centering
        \includegraphics[width=\linewidth]{figs/ohi_scores_globalnorm_12.png}
        \caption{7 selected heads $K_{sel}=28$.}
    \end{subfigure}
    \vspace{-0.05in}
    \caption{\textbf{Head selection patterns under different head budgets.} The head selection procedure uses default hyper-parameters of $\tau = 0.5$ and $\rho = 2$. All results are obtained on ScreenSpot-Pro using the 1:2 input-text:image token ratio.}
    \label{fig:headselection_different_tau}
\end{figure*}

%% file: figs/headselection_7B_different_ratios.tex
\begin{figure*}[t]
    \centering
    \begin{subfigure}{0.28\textwidth}
        \centering
        \includegraphics[width=\linewidth]{figs/ohi_scores_globalnorm_12.png}
        \caption{Head selection for Qwen2.5-VL-7B under a input-text:image token ratio of 1:2.}
    \end{subfigure}
    \quad
    \begin{subfigure}{0.28\textwidth}
        \centering
        \includegraphics[width=\linewidth]{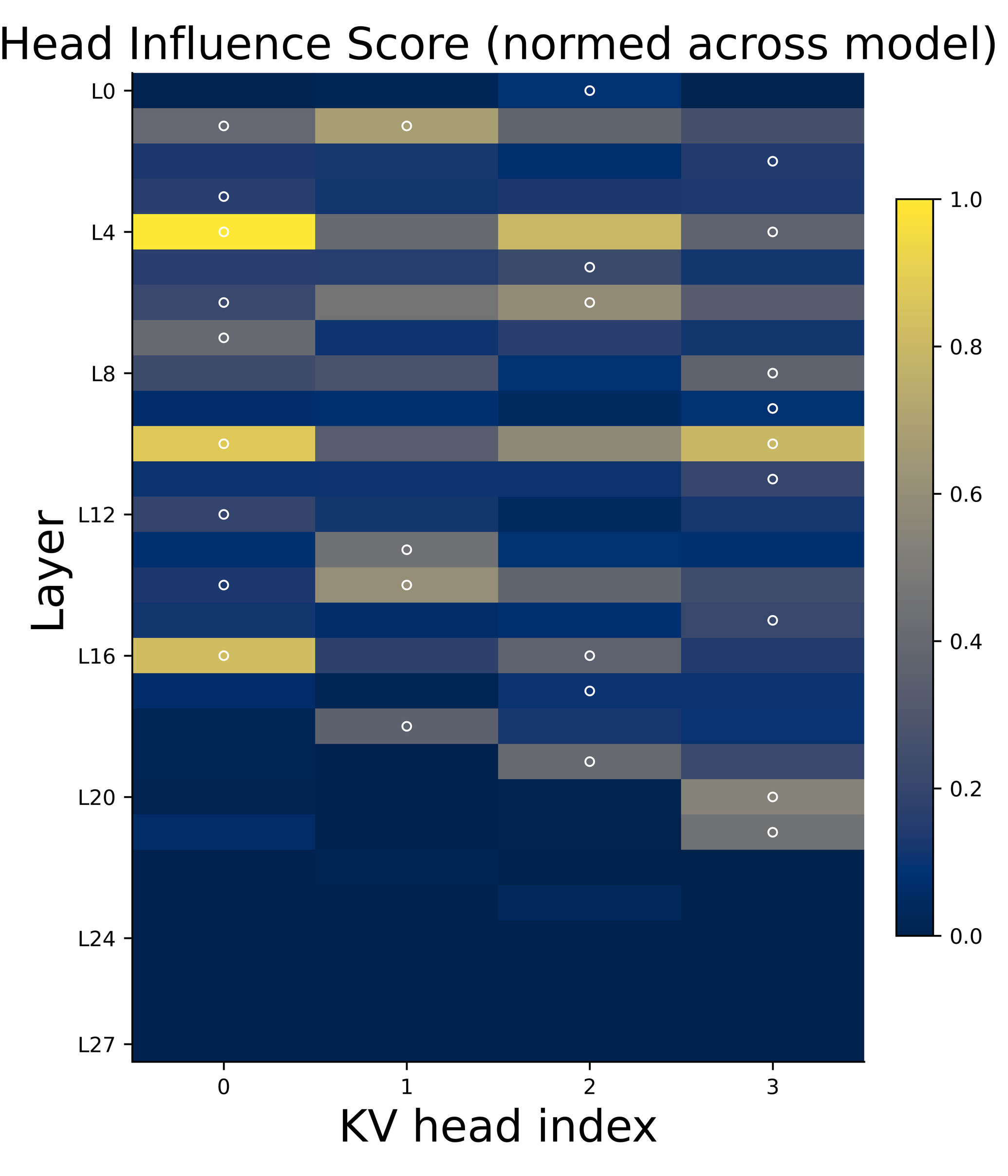}
        \caption{Head selection for Qwen2.5-VL-7B under a input-text:image token ratio of 1:3.}
    \end{subfigure}

    \medskip

    \begin{subfigure}{0.28\textwidth}
        \centering
        \includegraphics[width=\linewidth]{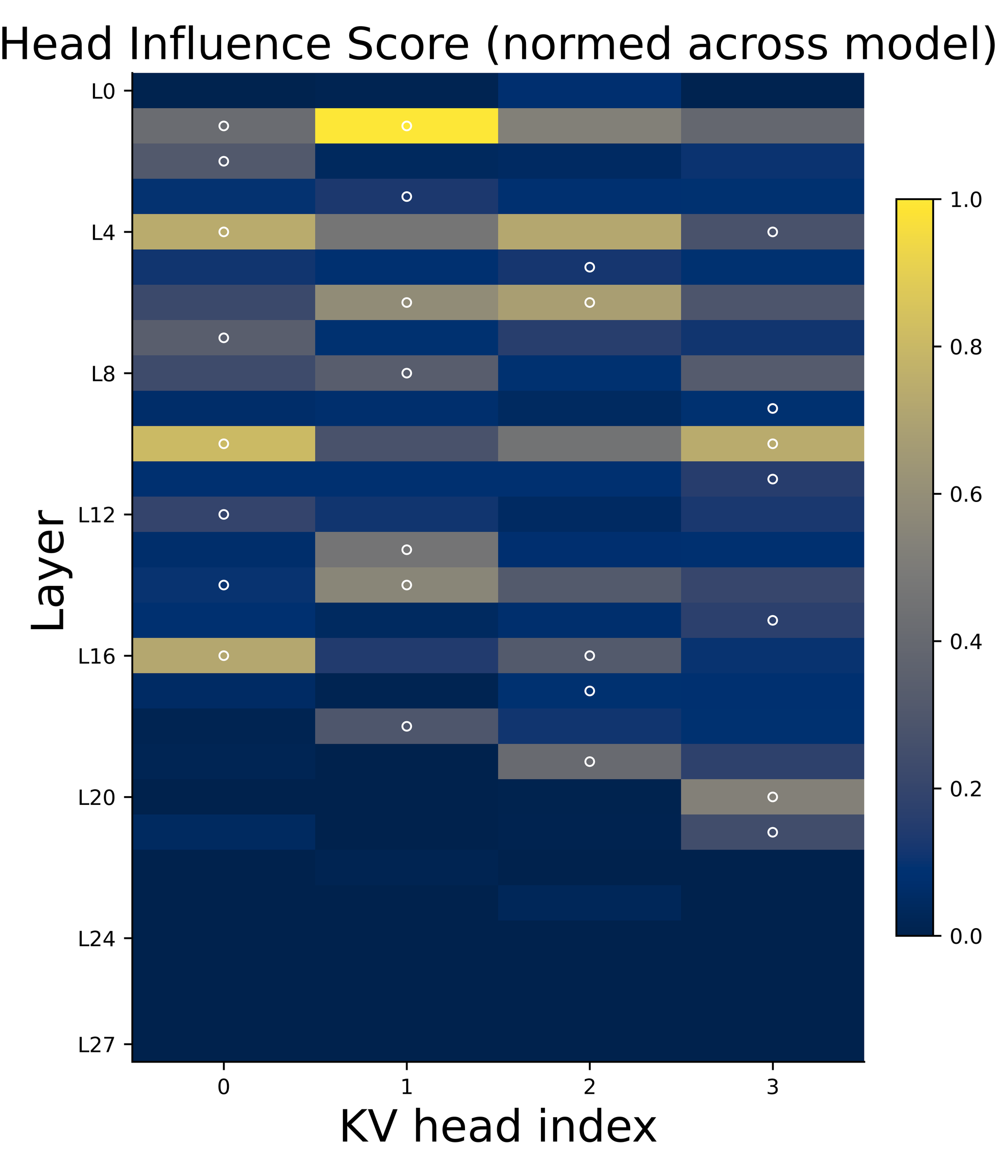}
        \caption{Head selection for Qwen2.5-VL-7B under a input-text:image token ratio of 1:4.}
    \end{subfigure}
    \quad
    \begin{subfigure}{0.28\textwidth}
        \centering
        \includegraphics[width=\linewidth]{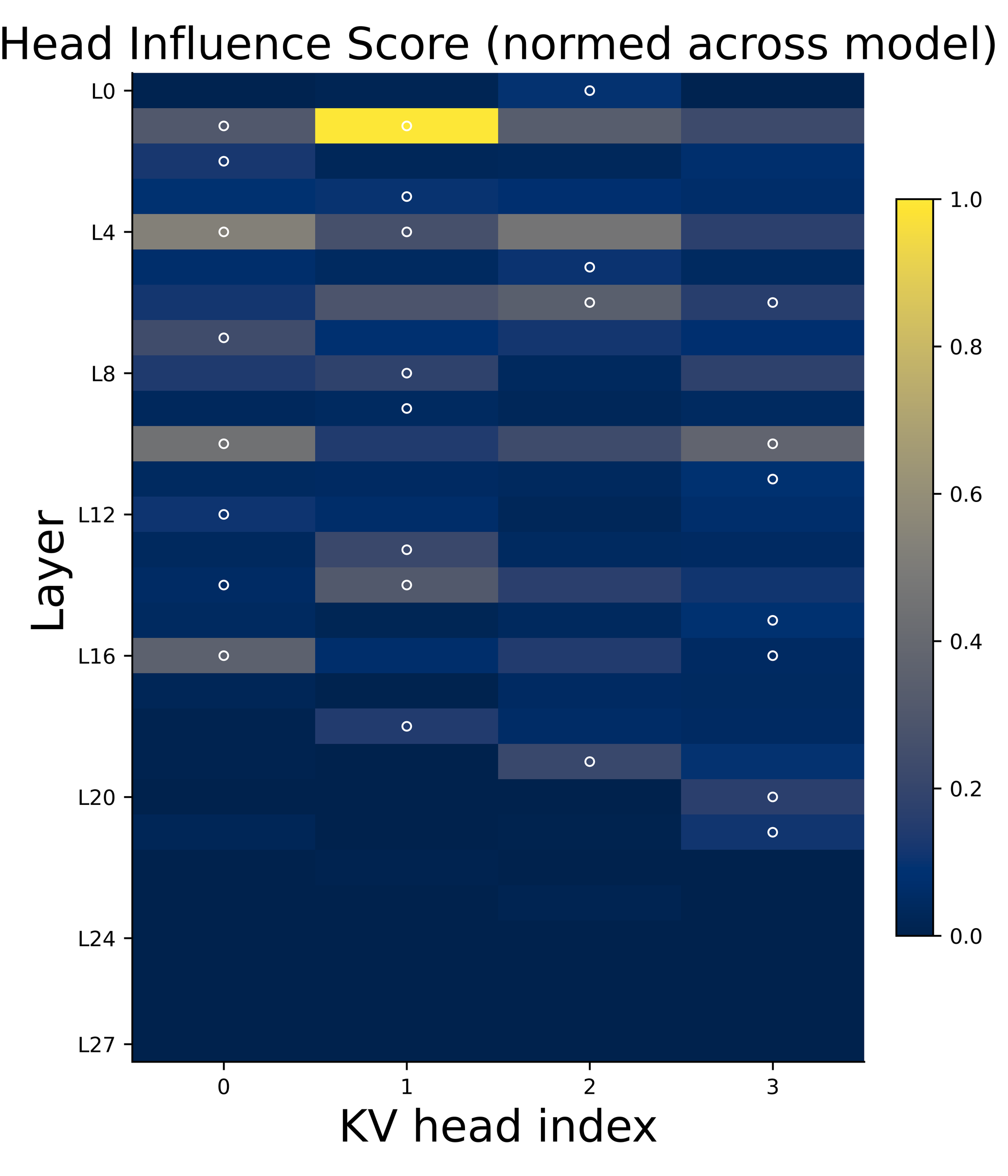}
        \caption{Head selection for Qwen2.5-VL-7B under a input-text:image token ratio of 1:5.}
    \end{subfigure}

    \vspace{-0.05in}
    \caption{\textbf{Head selection for Qwen2.5-VL-7B across different input-text:image token ratios on ScreenSpot-Pro.} Although ratios (and thus image resolutions) differ, the resulting head selections remain similar, with minor variations.}
    \label{fig:headselection_7B_different_ratios}
\end{figure*}